\newcounter{myctr}
\def\myitem{\refstepcounter{myctr}\bibfont\parindent0pt\hangindent13pt\themyctr.\enskip}

\documentclass{ws-ijhr}

\usepackage{algorithm}
\usepackage{chngcntr}
\usepackage{amssymb}
\usepackage{mathtools}
\usepackage{algorithmic}
\usepackage{subfigure}
\usepackage{graphicx}
\usepackage{color}

\begin{document}
	
	\markboth{Yan WEI, Wei JIANG, Ahmed RAHMANI, Qiang ZHAN}{Motion Planning for a Humanoid Mobile Manipulator System}
	
	%
	%
	
	\title{Motion Planning for a Humanoid Mobile Manipulator System} 
	
	\author{Yan Wei, Wei Jiang\footnote{Corresponding author.}, Ahmed Rahmani} 
	
	\address{CRIStAL, UMR CNRS 9189, Ecole Centrale de Lille\\ Villeneuve d'Ascq, 59650, France\\ 
		wytt0324@gmail.com, wjiang.lab@gmail.com}
	
	\author{Qiang Zhan}
	
	\address{Robot Research Institute, Beihang University\\
		Beijing, 100191,  China}

	\maketitle
	
	
	\begin{abstract}
		A high redundant non-holonomic humanoid mobile dual-arm manipulator system (MDAMS)  is presented  in this paper where the motion planning to realize ``human-like'' autonomous navigation and manipulation tasks is studied. 
		Firstly, an improved MaxiMin NSGA-II algorithm, which optimizes five objective functions  to solve the problems of singularity, redundancy, and coupling between mobile base and manipulator simultaneously,
		is proposed to design the optimal  pose to manipulate the target object.
		Then, in order to link the initial pose and that optimal pose, an off-line motion planning algorithm is designed. In detail, an efficient direct-connect bidirectional RRT and gradient descent algorithm  is proposed  to reduce the sampled nodes largely, and a geometric optimization method is proposed for path pruning.
		Besides,  head forward behaviors are realized by calculating the reasonable orientations and assigning them to the mobile base to improve the quality of human-robot interaction.  
		Thirdly,  the extension to on-line planning is done by introducing real-time sensing, collision-test and control cycles to update robotic motion in dynamic environments. 
		Fourthly, an EEs' via-point-based multi-objective genetic algorithm (MOGA) is proposed to design the ``human-like'' via-poses  by optimizing four objective functions.
		Finally, numerous simulations are presented to validate the effectiveness of  proposed algorithms.
	\end{abstract}
	

\section{Introduction}
\noindent
Industrial manipulators have been widely used to do some repetitive tasks, but they are usually fixed  manipulators in structured environments with predefined task policies and control inputs.  
Recently, more and more mobile manipulators are utilized in  fields from personal assistance to military applications$^{1-5}$. 
However, personal robots, normally humanoid mobile manipulators, are required to  work automatically  in unstructured environments with  humans. 
In order to improve the quality of  human-robot  interaction, robots' behaviors should be predictable and natural. 
The existing personal assistant, entertainment, accompany or guidance robots (e.g., sweeping robots, \emph{AMI}$^{1}$ and \emph{Pepper}$^{2}$) 
are usually mono-functional,  non-programmable, non-interactive or too specialized for  non-expert users. 
Therefore, the research of multi-functional, completely autonomous and sustainable personal robots  is still open and challenging.$^{5,6}$  

Many  motion planning methods$^{7-10}$ 
for multiple degrees of freedom (DoFs) robots have been presented, but they are incapable of solving motion planning problems under multiple constraints. 
Direct and inverse kinematics (IK)-based indirect  motion planning methods for  multi-DoFs manipulators$^{11-19}$ 
have been proposed. 
However, for direct  methods, the EE's motion is difficult to predict. For IK-based indirect methods, though EE's motion is guaranteed, they need to optimize a predefined objective function which must be continuous$^{17-19}$, 
and require the inverse Jacobian calculation which leads to singularity problem and results in unpredictable behaviors due to the high non-linearity between the task and joint spaces.
In order to overcome these shortcomings,  evolutionary algorithms represented by genetic algorithms (GAs) are used$^{20-25}$ 
since GA 
has no constraints on  cost functions (they can be discontinuous, not differentiable, stochastic, or highly nonlinear), no need of calculating inverse Jacobian, and is capable of optimizing (maximizing and minimizing simultaneously) multiple objectives, etc.$^{26}$
In reality,  the multiple objectives  usually conflict with each other.
By defining a combined fitness function, MOGAs can also take various objectives into account at the same time. For instance, the multi-objectives are EE's positioning error  and joint displacement for a redundant  manipulator$^{22}$ 
and for a mobile manipulator$^{24}$, 
mobile base and joint's displacements  for a mobile manipulator$^{23}$, 
the navigation length, path-obstacles intersection and accumulated change of mobile base's orientation  for a mobile vehicle$^{25}$,  
etc.
However, most of them do not study the coordination between the position-orientation of  mobile base and manipulator's configuration, and not mention the ``human-like'' motion design. 

On the other hand, naturally, the presence of multiple objectives can result in a set of optimal solutions (known as Pareto-optimal solutions).
The problem is that the combined fitness function-based MOGAs  can only generate one optimal solution and needs to specify the weighting coefficients  in advance, which means that this algorithm loses the diversity.
In order to preserve the diversity, Srinivas$^{27}$ 
proposed a non-dominated sorting genetic algorithm (NSGA) searching for the optimal solution in multiple directions. As a result, multiple solutions are generalized to form a Pareto-optimal set. 
However,  NSGA  needs to specify a predefined sharing parameter. To release this constraint, Deb$^{28}$ 
proposed the NSGA-II by introducing a fast-non-dominated-sort scheme and a crowding distance assignment. In the above two works, Pareto-optimal solutions in each generation are not well distributed. Hence, Pires$^{29}$ 
introduced the MaxiMin sorting scheme into NSGA-II where only the last selected front is sorted, but the previous fronts are not. This is why we propose an improved MaxiMin NSGA-II algorithm in which  not only the last selected front but all the selected fronts are sorted using the MaxiMin sorting scheme. 
In this way, the probability of inheriting good genes is improved as much as possible at each generation.

Another problem is that since the off-line designed motion may fail by colliding with  obstacles in dynamic environments,
on-line motion planning methods, 
especially  RRTs-based methods, attract great interest due to their high efficiency.$^{30-32}$ 
As many motion planning methods generate low quality paths, 
path pruning techniques  are proposed based on the medial axis.$^{33,34}$ However, they  require calculating the medial axis in the collision-free configuration space $C_{free}$ which is complex. 
The RRT*$^{31}$ 
consists of \emph{reconnect} and \emph{regrow} processes and  only guarantees local optimal path. 
The Informed-RRT*$^{32}$ 
is asymptotically optimal but it needs to evaluate the  whole  path which is time-consuming and unnecessary.
Besides, they 
mainly focus on  motion planning for mobile or free-flying robots and do not consider the non-holonomic constraints.$^{31,32}$ 
Even thought there exist studies on non-holonomic mobile vehicles$^{35}$, 
they do not take into account the heading angle which is very important for non-holonomic humanoid mobile manipulators studied in this paper. 

Theoretically speaking, the above mentioned works 
can be applied to  high DoFs robots, but their effectiveness will decrease and they are incompetent to solve multiple constrained problems.$^{30-35}$
Many researchers work on motion planning for multi-DoFs  manipulators$^{36,42}$ 
in real time. 
Lee$^{36}$ 
proposed a virtual roadmap-based  on-line trajectory generation method for a dual-arm robot.  
Berg$^{37}$ 
planned the motion in state-time space using roadmap and A*. 
Singularity avoidance of a mobile surgery assistant was studied$^{37}$ 
via a penalty function, but the EE's path is needed in advance. 
Yang$^{39}$ 
introduced a configuration re-planning  method for a fixed manipulator in dynamic environments. However, it requires calculating  inverse Jacobian and the desired configuration must be known.
Chen$^{40}$ 
designed a joint velocity correction term to the manipulator's joint trajectory. 
Xin$^{41}$ 
introduced an escape velocity and projected it onto the null space for redundant arms. 
Han$^{42}$ 
used the distance calculation and discrete detection for robot arms.
However, they do not deal with the multiple constrains or ``human-like'' behaviors.
In order to realize ``human-like'' behaviors,
a heuristic method was presented for a multi-DoFs humanoid robot arm using the RRT* algorithm directly in task space.$^{43}$ 
However, only the EE and elbow are considered as two hierarchical control points. In fact, as the number of control points increases, the hierarchical motion planning method will become much more complicated. 
Vannoy$^{44}$ 
generated a population of trajectories based on the fitness evaluation for  mobile manipulators in dynamic environments with unknown moving obstacles; however, this method is time-consuming.
Zhao$^{45}$ proposed a motion-decision algorithm to realize ``human-like'' movements only for a redundant arm.
Besides, many reasoning methods, like analytic  and fuzzy logic methods, 
have been introduced to achieve ``human-like'' behaviors.$^{46-49}$  
However, the specific consideration and  rules for different scenarios are required  which are complicated for high DoFs systems.

To summarize, in the above-mentioned methods, the desired pose or the EE's path needs to be known in advance. However, they can be unknown in this paper. The  ``human-like''   behavior design for the humanoid robot is still an open problem.
In this paper, only the EEs' desired positions-orientations are known. The \emph{optimal pose} (see \textbf{Definition 1}) and the motions in task and joint space are to be designed. The objective is to make the MDAMS be  ``human-like''  and  ``natural''  as much as possible in both their appearance and behaviors to guarantee the quality of human-robot interaction.  
The main contributions are listed as follows:
\begin{itemlist}
	\item An improved MaxiMin NSGA-II  \textbf{Algorithm \ref{alg:MaxiMinNSGA}}  is proposed in Section \ref{sec:mmnsga}. It is used to design the \emph{optimal pose} given only EEs' desired positions-orientations. Five objective functions are  optimized simultaneously to achieve  ``natural''  behaviors. The position-orientation of mobile base and the configuration of manipulator are planned at the same time.
	\item A fast off-line motion planning \textbf{Algorithm \ref{alg:offline}}  is given in Section \ref{sec:3} considering the \emph{optimal pose} has been designed by \textbf{Algorithm \ref{alg:MaxiMinNSGA}}. 
	\item A direct-connect  bidirectional RRT and gradient descent sampling process is proposed to improve the performance of RRTs in Section \ref{sec:offalg}. 
	\item An efficient geometric optimization method pruning the sampled path via node rejection and node adjustment, is proposed  to always guarantee  the shortest consistent path for repeated tasks in Section \ref{sec:gomoptm}.
	\item An on-line motion planning \textbf{Algorithm \ref{alg:online}} is designed  in Section \ref{sec:4}  by introducing on-line sensing, collision-test and control cycles  with the dynamic obstacles being treated in real time. In order to take the via-poses into consideration, an EEs' via-point-based  MOGA  algorithm is presented. Four objective functions are defined to optimize the candidate via-poses  corresponding to the EEs' via-points.
\end{itemlist}
Besides, the forward kinematics for  MDAMS is established  using the modified Denavit-Hartenberg (MDH) method in Section \ref{sec:2}, and a number of motion planning simulations are presented in Section \ref{sec:5}.

\section{System  Modeling of MDAMS}\label{sec:2}
\begin{figure}[th]
	\centerline{{\scriptsize (a)}\psfig{file=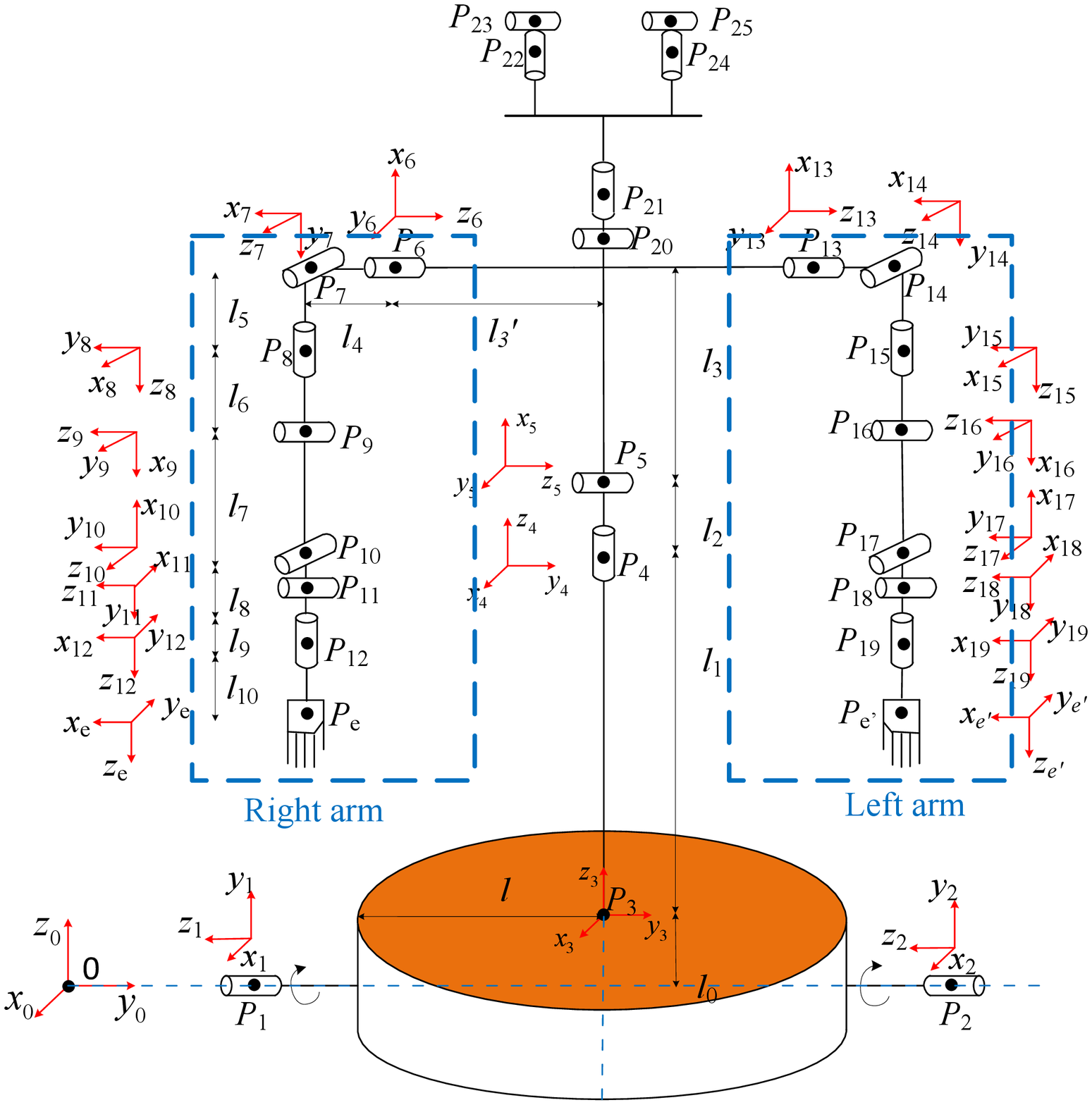,width=7cm}{\scriptsize (b)}\psfig{file=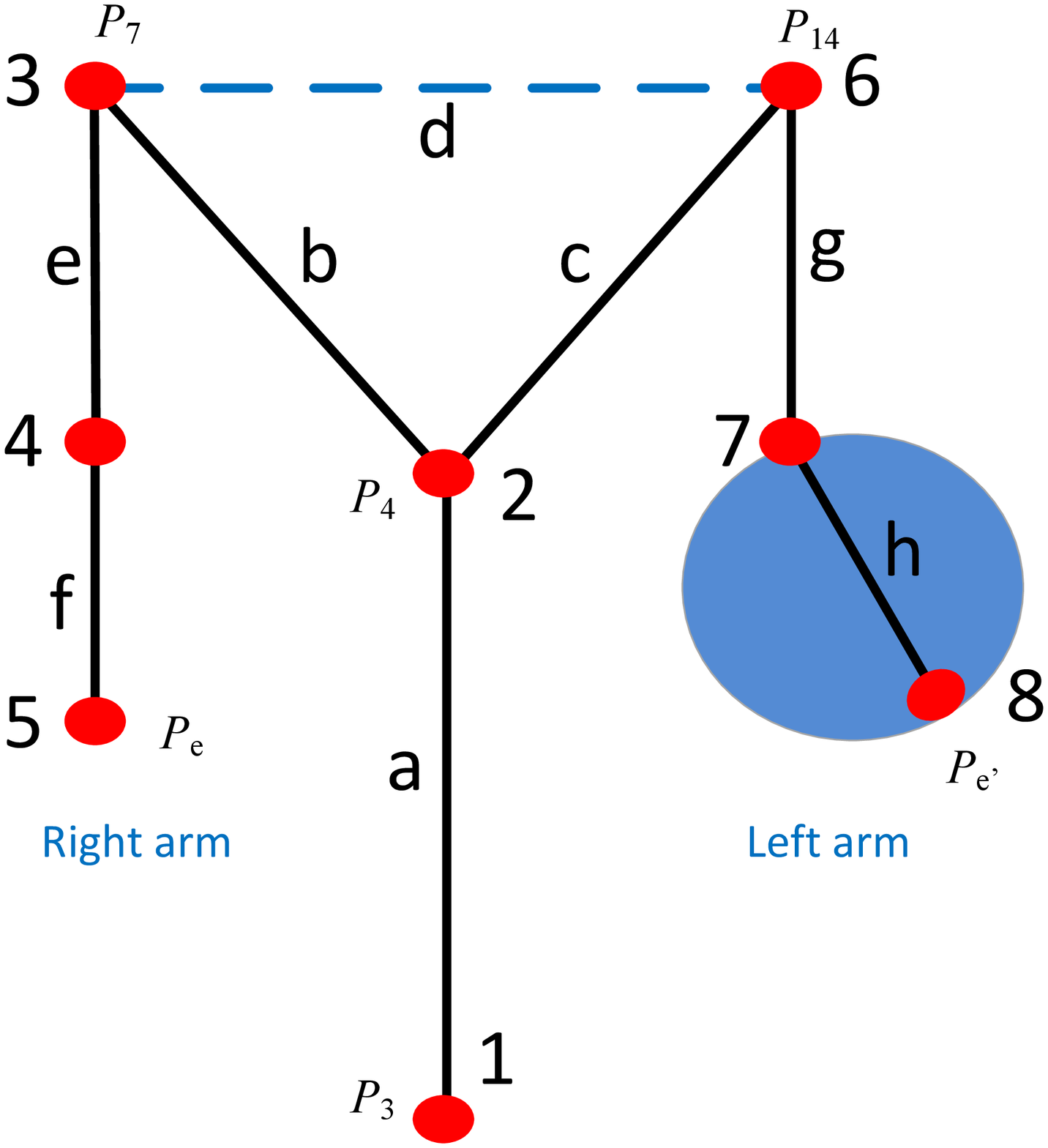,width=5cm}} 
	\vspace*{8pt}
	\caption{Mechanical description of MDAMS. (a) System structure; (b) System simplification.}
	\label{fig:robotmodel}
\end{figure}

\noindent
The structure of the designed MDAMS  is illustrated in Fig. \ref{fig:robotmodel} (a).
It has two arm-hand subsystems sharing  one common waist  and one common non-holonomic mobile base.  
The MDH method  is used to characterize the motion of frame $\Sigma_i$ in  $\Sigma_{i-1}$ via the transformation matrix: 
\begin{equation}
T_{i-1}^i = Rot(x,\alpha_{i-1})  Trans(x,a_{i-1})  Rot(z,\theta_i)  Trans(z,d_i)= \left[ \begin{array}{cc}
R_{i-1}^i & P_{i-1}^i\\
0 & 1 \end{array} \right] 
\end{equation}
through a sequence of rotations $ Rot() $ and translations $ Trans() $  with only four parameters  $ \alpha_{i-1} $, $a_{i-1}$, $\theta_i$ and $d_i$.  $ R_{i-1}^i $ and $ P_{i-1}^i $ are  the rotation matrix and position vector of  $ \Sigma_i $ in  $ \Sigma_{i-1} $, respectively.

Apply the MDH method to the designed MDAMS, $ 22 $ frames are defined without  considering  the neck, head and fingers$^{50}$. 
Specifically,  $  \Sigma_0 $ is the world frame;  $ \Sigma_i $ $(i=1,...,19)$ is the frame with its origin locating at point $P_i$; $P_e$ and $P_{e'}$ are the palm centers (EEs); $\Sigma_{e}$ and $\Sigma_{e'}$ are  EEs' frames  with a translation along  axis $z_{12}$ and $z_{19}$, respectively.  
In particular,   frame $\Sigma_3$ is attached to the mobile base  which is described by  $\bm{\theta}_m=(x, y, \phi) $ where $ (x, y) $ and $ \phi $ are mobile base's position and orientation in  $ \Sigma_0 $, respectively. 
$ l_0 $ is the translation of  $\Sigma_{3}$ along axis $z_0$ of  $ \Sigma_0 $ and $ l_k $ $(k=1,...,10)$  characterizes the  geometric dimension (see Table $ 1 $) of MDAMS in which the variables are defined as follows:

\begin{itemlist}
	\item $\bm{\theta} \overset{\Delta}{=} \left[\begin{array}{cccc}
	\bm{\theta}_m^T & \bm{\theta}_\omega^T & \bm{\theta}_R^T & \bm{\theta}_L^T
	\end{array}  \right]^T \in \mathbb{R}^n$: generalized variable,
	\item $T_{R} = T_{0}^3 T_{3}^4 T_{4}^5 T_{5}^{6}T_{6}^{e} (\bm{\theta}_m,\bm{\theta}_w, \bm{\theta}_R)$:  transformation matrix of the right EE in $\Sigma_0$,
	\item  $T_{L} = T_{0}^3 T_{3}^4 T_{4}^5 T_{5}^{13} T_{13}^{e'} (\bm{\theta}_m, \bm{\theta}_w, \bm{\theta}_L)$:  transformation matrix of the left EE in $\Sigma_0$,
	\item $T_{Rb} =  T_{3}^4 T_{4}^5 T_{5}^{6} T_{6}^{e} (\bm{\theta}_w, \bm{\theta}_R)$:  transformation matrix of the right EE in  $\Sigma_3$,
	\item  $T_{Lb} = T_{3}^4 T_{4}^5 T_{5}^{13} T_{13}^{e'} (\bm{\theta}_w, \bm{\theta}_L)$:  transformation matrix of the left EE in $\Sigma_3$,
	\item $\dot{x}_R = \left[\begin{array}{cc} 
	J_{v_R} & \bm{0}_{n_{L}} 
	\end{array}\right] \dot{\bm{\theta}}= J_{R} \dot{\bm{\theta}} \in \mathbb{R}^m$: linear velocity of the right EE in $\Sigma_0$, 
	\item  $\dot{x}_L = \left[\begin{array}{ccc}
	J_{v_{Lo}} & \bm{0}_{n_{R}} & J_{v_{La}}
	\end{array}\right] \dot{\bm{\theta}}$ $= J_{L} \dot{\bm{\theta}}  \in \mathbb{R}^m$: linear velocity of the left EE  in $\Sigma_0$,
	\item $\omega=\left[\begin{array}{cc}
	\omega_R^T&\omega_L^T
	\end{array}\right]^T = \left[\begin{array}{cc}
	J_{\omega_R}^T & J_{\omega_L}^T 
	\end{array}\right]^T \dot{\bm{\theta}}= J_{\omega}\dot{\bm{\theta}}  \in \mathbb{R}^{2m}$: EEs'  angular velocities in $\Sigma_0$,
\end{itemlist}
where $\bm{\theta}_w \in \mathbb{R}^{n_w} $, $\bm{\theta}_R \in \mathbb{R}^{n_R} $ and $\bm{\theta}_L \in \mathbb{R}^{n_L} $ represent  the  joints  in  waist,  right and left arms, respectively. $(\bm{\theta}_w, \bm{\theta}_R, \bm{\theta}_L)$ represents the configuration of  manipulator.
$J_{v_R} \in \mathbb{R}^{m\times (n_m+n_w+n_R)}$ is the    right EE's Jacobian in  $\Sigma_0$; 
$J_{v_{Lo}}\in \mathbb{R}^{m\times (n_m+n_w)}$ and $J_{v_{La}}\in \mathbb{R}^{m\times n_R}$ are the left EE's Jacobians in  $\Sigma_0$;  
$J_{\omega_i}$ $(i = R,L)$ is the EE's orientation Jacobian. 
In this paper, $n_m = 3$, $n_w=2$, $n_R=n_L=7$,  $n=n_m+n_w+n_R+n_L=19$ and $m=3$.

\begin{table}[h]
	\centering
	\caption{Robot's dimension.}
	\label{tbl:robDim}
	\begin{tabular}{l|l|l|l|l|l|l|l|l|l|l|l}
		\hline
		l    & $l_0$ & $l_1$ & $l_2$ & $l_3$/ $l_{3}'$ & $l_4$ & $l_5$ & $l_6$ & $l_7$ & $l_8$ & $l_9$ & $l_{10}$ \\ \hline
		0.16 & 0.262 & 0.45  & 0.15  & 0.3   & 0.15  & 0.2   & 0.05  & 0.2   & 0.05  & 0.05  & 0.05  \\ \hline  
	\end{tabular}
\end{table}

\begin{definition}[{Pose} and {Optimal pose}]\label{defn:pose}
	Throughout this paper, we note \emph{pose} as the position-orientation of mobile base  and the configuration of upper manipulator (i.e. $\bm{\theta}$) for the designed MDAMS. $\bm{\theta}_{op}$ denotes the \emph{optimal pose} which reaches EEs' desired positions-orientations $\bm{X}_{EE}$  under multiple constraints.
\end{definition}

Suppose that the initial \emph{pose}  and  $\bm{X}_{EE}$  of  MDAMS are known. 
In the rest of this paper, {optimal pose} design, off-line and on-line motion planning problems for MDAMS are investigated.

\section{Optimal Pose Design}\label{sec:mmnsga}
\noindent
The GA is a popular method for solving optimization problems based on  natural genetics and selection mechanics$^{51}$. 
A GA drives a population which is composed of many individuals that are usually represented by a set of parameters (known as chromosomes), to evolve under specified selection rules to a state that maximizes the fitness function.   The population of individuals (possible solutions) which is initialized randomly, is modified repeatedly through selection, crossover and mutation operators. At each step, the GA selects good individuals to be parents from the current population, and uses them to produce the offspring for the next generation (crossover). Random changes are introduced to the population by means of mutation operator.  The population evolves towards an optimal solution over successive generations. 

\subsection{Objective functions}\label{Sec:Objfunc}
\noindent
Due to the high redundancy, there exist numerous joints combinations given  EEs' desired  positions-orientations $\bm{X}_{EE}$, and there is always a preference.
Instead of only taking into account  EE's positioning accuracy or joint displacement, five objective functions are considered simultaneously as follows:

\begin{itemlist}
	\item \textbf{EE's positioning accuracy}: $f_1(\bm{\theta}) = \|x_R(\bm{\theta}) - x_{Rd}\| + \|x_L(\bm{\theta}) - x_{Ld}\|$, where $x_{Rd},\, x_{Ld}$, $x_R(\bm{\theta})$ and $ x_L(\bm{\theta})$ are  the desired and real positions of  right and left EEs, respectively.
	\item \textbf{EE's orientation tracking accuracy}: $f_2(\bm{\theta}) = \| e_{q_R}(\bm{\theta})\| + \| e_{q_L}(\bm{\theta})\|$, where $ e_{q_R}(\bm{\theta}) $ and $ e_{q_L}(\bm{\theta}) $ are respectively the orientation tracking errors of right and left EEs using the unit quaterion representation.$^{52}$ 
	\item \textbf{EE's manipulability}:
	Apart from reaching $\bm{X}_{EE}$, the additional criteria is required to achieve the ``human-like'' behaviors.  
	The manipulability ability describes the distance away from robot's singular configuration. 
	The usually used criterion is  $\Omega = \sqrt{det(J(\bm{\theta}) J^T(\bm{\theta}))}$ or $\Omega=\frac{\sigma_{max}(J(\bm{\theta}))}{\sigma_{min}(J(\bm{\theta}))}$ where $ J(\bm{\theta}) $ is EE's Jacobian and $\sigma$ is the singular value of $J(\bm{\theta})$.
	Then, the following objective function is given:\\
	$f_3(\bm{\theta}) = -(\Omega_{R}(\bm{\theta})+\Omega_{L}(\bm{\theta}))$, where $ \Omega_{R}(\bm{\theta}) $ and $ \Omega_{L}(\bm{\theta}) $ are the manipulability measures of  right and left EEs, respectively.
	\item \textbf{Joint displacement}:
	In order to save energy, the least joint displacement is required. 
	A mass-based joint displacement objective function is proposed  as follows:  \\
	$f_4(\bm{\theta}) = \frac{1}{n}\sum_{i=1}^{n} W_i\frac{\bm{\theta}_i - \bm{\theta}_{imin}}{\bm{\theta}_{imax} - \bm{\theta}_{imin}}$, where  $n$ is the dimension of the generalized variable $\bm{\theta}$,  $\bm{\theta}_i$ is the $i$-th generalized variable, and $\bm{\theta}_{imax}$ and $ \bm{\theta}_{imin}$ are the corresponding boundaries. $W_i={M_i}/{ \sum_{j=1}^{{n}} M_j}$ is  mass-based weight,  where $M_i= \sum_{k=i}^{n_{e}} m_k $,  $m_k$ is the mass of link $k$, and $n_{e}$ is the index of EE (left or right). $f_4(\bm{\theta})$ means that the most heavy part has the largest difficulty to move.
	\item \textbf{EE's displacement w.r.t. the mobile base}:
	Since the above four functions mainly focus  on the manipulator, the coordination between  mobile base and manipulator is not studied.
	With regard to this, EEs'  displacements in the frame  $\Sigma_{3}$ of mobile base  is \\
	$f_{5}(\bm{\theta}) = \|x_{Rb}(\bm{\theta})- x_{Rbi}\| + \|x_{Lb}(\bm{\theta}) - x_{Lbi}\|$,  where $x_{Rbi} $  and $ x_{Rb}(\bm{\theta})$ ($x_{Lbi} $  and $ x_{Lb}(\bm{\theta})$) are  the initial and desired  positions  of  right (left) EE in  $\Sigma_{3}$, respectively.
\end{itemlist}

\subsection{Improved MaxiMin NSGA-II algorithm}
\noindent
In this section, the improved  MaxiMin NSGA-II  \textbf{Algorithm \ref{alg:MaxiMinNSGA}}  is proposed to design the optimal pose $\bm{\theta}_{op}$ by optimizing the above five objective functions. 
The chromosome  is chosen as $\bm{\theta}=(x,\, y,\,\phi,\, \bm{\theta}_{\omega},\,\bm{\theta}_{R},\,\bm{\theta}_{L})$ and its searching interval locates in  $C_{free}$. 

\begin{figure}[b]
	\centerline{\psfig{file=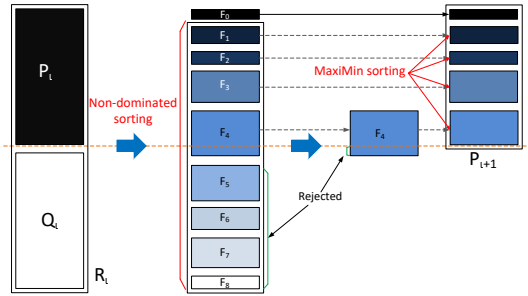,width=7cm}} 
	\vspace*{8pt}
	\caption{Flowchart of the improved MaxiMin NSGA-II algorithm.}
	\label{fig:nsga2}
\end{figure}

\begin{algorithm}[h]
	\caption{Improved MaxiMin NSGA-II}\label{alg:MaxiMinNSGA} 
	\begin{algorithmic}[1]
		\STATE \textit{Initialization} : rand($P_0$) \COMMENT{initialize the population randomly}
		\STATE $sort(P_0) $ \COMMENT{sort $P_0$ according to Eq. (\ref{eq:fitdis})}
		\STATE $Q_0=BasicGA(P_0) $ \COMMENT{usual GA reproduction}
		\\ $ \iota=0 $ \COMMENT{initialize the generation numeration}
		\WHILE{$ \iota\leq N_G $}
		\STATE $R_\iota = P_\iota\cup Q_\iota$ \COMMENT{merge population to size $ 2N_{pop} $}
		\STATE  $F_0=\phi$ \COMMENT{initialize selected set $F_0$}
		\FOR {$ i=1$  \TO $ N_{obj} $}  
		\STATE $F_0 = F_0\cup GetMin(R_\iota,\,i)$ \COMMENT{move the individual with the minimum  objective function $f_i$ from  $R_\iota$ to $ F_0 $}
		\ENDFOR
		\STATE $P_{\iota+1} = F_0$ \COMMENT{initialize the parent population $ P_{\iota+1} $ for the next generation}
		\STATE $fast$-$no$n-$dominated$-$sort (R_\iota)$ \COMMENT{non-dominated sorting}
		\STATE $ i=1 $ \COMMENT{get the first non-dominated set $F_i$ in the sorted population $R_\iota$}
		\WHILE{$\#P_{\iota+1}+\#F_i\leq N_{pop}$}
		\STATE $F_i = MaxiMin$-$sort(F_i)$ \COMMENT{sort $F_i$ using the MaxiMin sorting scheme}
		\STATE $ P_{\iota+1}=P_{\iota+1} \cup F_i $ \COMMENT{add $F_i$ to the population $ P_{\iota+1} $}
		\STATE $ i=i+1 $ \COMMENT{increment of the non-dominated set counter}
		\ENDWHILE\label{NSGA1while}
		\FOR {$ j=1$  \TO $ \#F_i $}
		\STATE $c_{p_j} $ =$ \min_{\forall p_k \in P_{\iota+1}} $ $\{||f_{p_j} - f_{p_k}||\}$ \COMMENT{calculate the minimum fitness distance $c_{p_j} $ and assign it to each individual in $F_i$} 
		\ENDFOR
		\WHILE{$\#P_{\iota+1}< N_{pop}$}		 
		\STATE $p=getMaxCi(F_i)$  \COMMENT{get the individual with the largest minimum fitness distance in $F_i$}
		\STATE $ P_{\iota+1}=P_{\iota+1} \bigcup p $ \COMMENT{add $p$ to the selected population $ P_{\iota+1} $}
		\FOR {$ l=1$ \TO $ \#F_i  $}
		\STATE $c_{p_l} = \min_{\forall p_k \in P_{\iota +1}} \{||f_{p_l} - f_{p_k}||,\, c_{p_l}\}$ \COMMENT{update the minimum fitness distance of each individual in $F_i$}
		\ENDFOR 
		\ENDWHILE \COMMENT{Complete $P_{\iota+1}$ selection using the MaxiMin sorting scheme}
		
		\STATE $Q_{\iota+1}$ = $make$-$new$-$pop(P_{\iota+1})$ \COMMENT{keep the individuals' orders in the selected population $P_{\iota+1}$ and use crossover and mutation operators to produce the next offspring population}
		\STATE $\iota=\iota+1$ \COMMENT{increment of the generation counter}
		\ENDWHILE \label{NSGA2while}
	\end{algorithmic}
\end{algorithm}

The whole process is illustrated in Fig. \ref{fig:nsga2} and  \textbf{Algorithm \ref{alg:MaxiMinNSGA}}. At the beginning, a parent population $P_0$ of size $N_{pop}$ is randomly created (line $ 1 $). 
Sort all the individuals $ \{G_i, \, i=1,...,N_{pop}\} $ in  $P_0$  according to their fitness values   (line $ 2 $, see Eq. (\ref{eq:fitdis})). 
The technique \emph{BasicGA}, which employs the usual binary tournament selection, recombination and mutation operators, is used to create the first offspring population $Q_0$ of size $N_{pop}$   (line $ 3 $). The generation numeration is initialized $\iota=0$.  

Then, a \textbf{while} loop is activated to evolve the Pareto-optimal solutions (lines $ 4 $ through $ 30 $). 
At the beginning of each generation $\iota$, a combined population $R_\iota = P_\iota \cup Q_\iota$ of size $ 2N_{pop} $ is formed (line $ 5 $). 	
In $ R_\iota $, get the individuals  which have the minimum value for each objective function (lines $ 6 $ through $ 9 $) and initialize the next parent population $P_{\iota+1}$ (line $10$). The rest individuals in $ R_\iota $ are sorted using the fast non-dominated sorting scheme  according to their non-domination$^{24, 28}$. 


Afterwards, select the best non-dominated individuals set by set, and sort each set using the MaxiMin sorting scheme. In particular, if the number of individuals in the selected population $P_{\iota+1}$ and in the following set $F_i$ is smaller than $ N_{pop} $ ($\#P_{\iota+1}+\#F_i\leq N_{pop}$), continue the selection (lines $ 13 $ through $ 17 $); if not ($\#P_{\iota+1}+\#F_i> N_{pop}$), 
select the individuals one by one using  the MaxiMin sorting scheme until obtaining $ N_{pop} $ selected individuals (loops \textbf{for} and \textbf{while}, lines $ 18 $ through $ 20 $ and lines $ 21 $ through $ 27 $). 
Finally, the next offspring population is produced while keeping the order of the selected population $P_{\iota+1}$ and go to the next generation ($\iota = \iota+1$, line $ 29 $).

The MaxiMin sorting scheme is described here.$^{29}$ 
Firstly, the distance  between each non-dominated individual  $ p_j \in F_i  \, (j=1,...,\#F_{i}) $ and the individuals already selected $p_k\in P_{\iota+1} \, (k=1,...,\#P_{\iota+1})$ is evaluated and save the minimum distance $ c_{p_j} $ for each individual $ p_j \in F_i$ (loop \textbf{for}, lines $ 18 $ through $ 20 $). 
Then, move the individual $ p_j \in F_i$ whose minimum fitness distance $c_{p_j}$ in  $F_i$ is maximum to $ P_{\iota+1} $ (lines $ 22 $ through $ 23 $). 
Every time an individual $p$ in $F_i$ enters in  $P_{\iota+1}$, the values $ \{c_{p_j}, \, j=1,...,\#F_{i}\} $ in $F_i$ are reevaluated (lines $ 24 $ through $ 26 $).

The main improvement of \textbf{Algorithm \ref{alg:MaxiMinNSGA}} is that the individuals in each non-dominated set $F_i$ are sorted using the MaxiMin sorting scheme  (line 14) compared with the MaxiMin NSGA-II algorithm in {\color{blue} Pires 05}.  
Therefore, good genes are inherited as much as possible by increasing the crossover possibility of good individuals. As a result, the converging speed of optimal solution is largely increased. 
What is more, \textbf{Algorithm \ref{alg:MaxiMinNSGA}} generates a Pareto-optimal set$^{24}$ 
instead of one single solution at each generation. 
And the following decision maker a posterior selects the optimal solution from the set at the end of each generation. 
\begin{equation}\label{eq:fitdis}
\underset{p_i \in P_{\iota +1}}{min} \, z_i = \sum_{j=1}^{n_{obj}} w_{ij} z_{ij}, \, i=1,...,N_{pop} 
\end{equation}
where $z_i$ is the fitness function of each individual $p_i \in P_{\iota +1}$, $w_{ij}$ is the weighting coefficient satisfying $\sum_{j=1}^{n_{obj}} w_{ij}=1$ and $ z_{ij} $ is the normalized $j-th$ objective function of individual $p_i$.

\section{Off-line Motion Planning}\label{sec:3}
\noindent
Given EEs' desired positions-orientations $\bm{X}_{EE}$, the optimal pose $\bm{\theta}_{op}$ can be designed using the   \textbf{Algorithm \ref{alg:MaxiMinNSGA}}. Similarly, given EEs' desired trajectories, the corresponding optimal motion of MDAMS can be designed,  but it will be time-consuming.
To this end,  an efficient point-to-point bidirectional RRT and gradient decent motion planning \textbf{Algorithm \ref{alg:offline}}  with a geometric optimization process is proposed in this section to design the motion for MDAMS given the optimal pose.

\subsection{Overview}\label{sec:offalg}
\begin{figure}[t]
	\centerline{\psfig{file=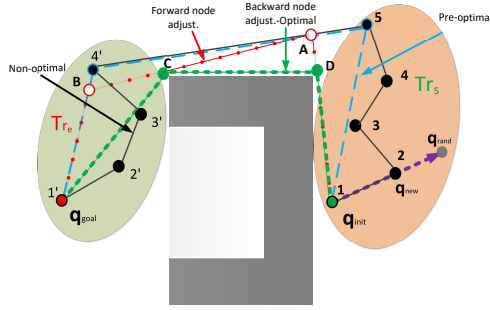,width=7cm}} 
	\caption{An example of off-line path planning.} 
	\label{fig:MBOptim}
\end{figure}

\noindent
Throughout \textbf{Algorithm \ref{alg:offline}}, the  path planner maintains two trees $Tr_s$  and $Tr_e$ which root respectively at the initial node $\bm{q}_{init}$ and final node $\bm{q}_{goal}$ (see Fig. {\ref{fig:MBOptim}}). 
A \textbf{while} loop is activated up till tree $Tr_s$  and tree $Tr_e$ meet each other, i.e. $  DisTree \le DisMax $  or there is a direct collision-free  connection between $Tr_s$  and  $Tr_e$. $DisTree$ is the minimum distance between $Tr_s$  and  $Tr_e$, $DisMax$ is a predefined value. 

\begin{algorithm}[h]
	\caption{Direct-connect BiRRT and Gradient descent Motion Planning}\label{alg:offline} 
	\begin{algorithmic}[1]
		\STATE \textit{Initialization} : start-node $ \leftarrow $ $\bm{q}_{init}$, end-node $ \leftarrow $ $\bm{q}_{goal}$
		\\ \STATE $Tr_s$  $ \leftarrow $  start-node, $Tr_e$  $ \leftarrow $  end-node  
		\WHILE{$ DisTree>DisMax $, or there is no collision-free direct connection}
		\STATE $ p_{rb}  \leftarrow  rand()$
		\IF{$ p_{rb}<p_{rbM} $}  
		\STATE {$[Tr_s, Tr_e] = BiRRT$-$Extend(Tr_s, Tr_e)$} 
		\ELSE
		\STATE {$[Tr_s, Tr_e] = GradDecExtend(Tr_s, Tr_e)$} 
		\ENDIF		
		\ENDWHILE
		\STATE $Path  \leftarrow  PathRRTGrad(Tr_s, Tr_e)$ \COMMENT{non-optimal path}
		\STATE $OptimPath  \leftarrow GeomOptim(Path)$ \COMMENT{geometric path optimization}
		\STATE $Motion  \leftarrow  InterPolyBlend(OptimPath, t_{start}, t_{end})$ \COMMENT{trajectory generation}
		\STATE \textbf{end}
	\end{algorithmic}
\end{algorithm}
During each \textbf{while} loop, the extending process $BiRRT$-$Extend$ (line $ 6 $) or $GradDecExtend$  (line $ 8 $) is selected (lines $ 3 $ through $ 10 $) to extend $Tr_e$ and  $Tr_s$ until two trees meet each other, based on the comparison between a randomly generated value $ p_{rb} $ (line $ 4 $)  and  a predefined value $ p_{rbM}  $ ($0 \leq p_{rbM} \leq 1$)  which is proportional to the clutter of environment. 
The purple dash line and nodes $\{\bm{q}_{init}$, $\bm{q}_{rand}$, $\bm{q}_{new}\}$ in Fig. \ref{fig:MBOptim} illustrate the conventional RRT extending process.$^8$ 
$BiRRT$-$Extend$ extends two trees at the same time following the above RRT process.
$GradDecExtend$ extends two trees directly along the linking line between two trees.
In particular, a direct-connect method is proposed even if the two trees $ Tr_s $  and  $ Tr_e $ are still very far away from each other. It links $ Tr_s $  and  $ Tr_e $ directly (segment $\bm{5}$-$\bm{4'}$) if there is a collision-free connection between two arbitrary nodes in $ Tr_s $  and  $ Tr_e $ respectively. As a result, the path searching process is greatly improved  by  reducing the sampled nodes.

Process $PathRRTGrad$ (line $ 11 $) finds the minimum path   from $\bm{q}_{init}$ to $\bm{q}_{goal}$ by linking two trees  $ Tr_s $  and  $ Tr_e $ as one single tree that roots at $\bm{q}_{init}$ and ends at $\bm{q}_{goal}$. 
Since the obtained path is normally non-optimal, process $GeomOptim$ (line $ 12 $) is proposed to  geometrically optimize the path  (see Subsection \ref{sec:gomoptm}). 
Finally, the process $InterPolyBlend$ (line $ 13 $, see Subsection \ref{sec:poly})  designs a time law  to  generate the time-specified trajectory within a desired time interval $[t_{start}, t_{end}]$. 

\subsection{Geometric path optimization: {$ GeomOptim $}}\label{sec:gomoptm}
\noindent
An example of the  path planning result is shown in Fig. \ref{fig:MBOptim}. The black line linking nodes $ \{\bm{1}(\bm{q}_{init}),$ $..., \bm{5}, \bm{4'},...,\bm{1'}(\bm{q}_{goal})\}$ represents the designed path after $BiRRT$-$Extend$ and $GradDecExtend$, which is obviously non-optimal. Therefore,  a geometric  optimization method which consists of the \emph{node rejection} and \emph{node adjustment} processes  is proposed. 
Note that the path after \emph{node rejection}  is  \emph{pre-optimal}, and the path after  \emph{node adjustment} is \emph{optimal}. The blue, red and green dash lines in  Fig. {\ref{fig:MBOptim}} illustrate the proposed method. 

\subsubsection{Node rejection} 
\noindent
The \emph{node rejection} process consists of the following steps: 
i)  initialize the \emph{pre-optimal path} with $ \bm{q}_{init} $; 
ii) test all the nodes along the black line in sequence, and reserve the last node $\bm{5}$ which does not interact with obstacles into the \emph{pre-optimal path};  
iii) repeat step ii) until reaching $ \bm{q}_{goal} $. 
The blue dash line which consists of nodes \{$ \bm{q}_{init} $, $ \bm{5} $, $\bm{4'} $, $ \bm{q}_{goal} $\} in Fig. \ref{fig:MBOptim} is the \emph{node rejection} result, i.e. the \emph{pre-optimal path}.

\subsubsection{Node adjustment}
\noindent
The obtained \emph{pre-optimal path} after  \emph{node rejection}  is usually non-optimal due to  the random sampling process. 
In order to further prune the path, a forward-backward  \emph{node adjustment} process is proposed as follows: 
i) introduce an \emph{auxiliary path} and initialize it  with $ \bm{q}_{init} $; 
ii) test in sequence the segments which link node $ \bm{q}_{init} $ and  the points on  segment $\bm{5} $-$ \bm{4'} $ (the point on segment $\bm{5} $-$ \bm{4'} $ moves from $\bm{5} $ to $\bm{4'} $  with a fixed step), and reserve the last node $ \bm{A} $ on segment $\bm{5} $-$\bm{4'} $  which does not collide with obstacles into the \emph{auxiliary path};  
iii) repeat  step ii) until reaching $ \bm{q}_{goal} $, then  a  path consisting of nodes \{$ \bm{q}_{init} $, $ \bm{A} $, $ \bm{B} $, $ \bm{q}_{goal} $\} is obtained;
iv) flip the obtained path and repeat steps i)-iii); 
v) flip the obtained path again and  the \emph{optimal path} containing  nodes \{$ \bm{q}_{init} $, $ \bm{D} $, $ \bm{C} $, $ \bm{q}_{goal} $\} is obtained. 
In Fig. \ref{fig:MBOptim}, the red dash line  represents the path after forward \emph{node adjustment}; the green dash line represents the path after backward \emph{node adjustment}  which is also the  \emph{optimal path} consisting of \emph{via-points} \{$ \bm{q}_{init} $, $ \bm{D} $, $ \bm{C} $, $ \bm{q}_{goal} $\}. 

\subsection{Interpolating linear polynomials  with parabolic blends}\label{sec:poly} 
\noindent
The \emph{optimal path} obtained after line  $ 12 $ in \textbf{Algorithm \ref{alg:offline}} is a geometric path, so a time law is needed.
On the other hand, the non-holonomic MDAMS is required to always head forward to show its movement intention to improve human-robot interaction.
Therefore, the process $InterPolyBlend$ is used (line 13)  to  design the smooth trajectory within time interval  $ [t_{start}, t_{end}] $ using the linear polynomials with parabolic blends interpolation technique$^{53}$. 
Consider the case where it is required to interpolate $ n_p $ via-points $ \{\bm{q}_{1}, ...,\bm{q}_{n_p}\} $ at time instants $\{ t_i, \; i=1,...,n_p, \; t_1=t_{start},\; t_{n_p}=t_{end}\} $, respectively. 
The  time scope between two adjacent via-points is 
\begin{equation}\label{eq:t_segment}
{\Delta t_k} = \frac{dis(\bm{q}_{k}, \bm{q}_{k+1})}{\sum_{i=1}^{n_p-1} dis(\bm{q}_{i}, \bm{q}_{i+1})} ({t_{end}-t_{start}}), \, k = 1,...,n_{p-1}.  
\end{equation}

In particular, the trajectory generation for the non-holonomic mobile base is described in the following. 
Suppose that the optimal  path of mobile base consists of $n_{p}$ via-points $\bm{q}_{k} = (x_k,y_k)$ $(k=1,\, 2,\,...\,,\, n_{p})$, $(x_1,y_1)$  and $(x_{n_{p}},y_{n_{p}})$ are respectively the initial and optimal positions, $ \phi_0$ and $\phi_{d}$ are the initial and optimal orientations, and  others are the intermediate ones. 
The orientation  $\phi_k$ between two adjacent via-points is calculated as follows: 
\begin{equation}
\phi_k = atan (\frac{y_{k+1}-y_k}{x_{k+1}-x_k}), \, k=1,...,n_{p}-1 
\end{equation} 
to always lead  MDAMS head forward.   As a result, the orientation list $(\phi_0, \phi_1,$ $...,\phi_{n_{p}-1}, \phi_d)$ is obtained. 
At each via-point $ \bm{q}_{k} $,  the mobile base firstly rotates to the heading orientation $\phi_k $,
then moves to the next via-point  $ \bm{q}_{k+1} $.

\section{On-line Motion Planning}\label{sec:4}   
\noindent
In the previous section, an efficient off-line motion planning algorithm is presented for MDAMS. 
However, the designed motion will fail if there are obstacle collisions.
Therefore, by extending \textbf{Algorithm \ref{alg:offline}}, an on-line motion planning \textbf{Algorithm \ref{alg:online}} inspired by {\color{blue} Mcleod 16} 
is introduced in this section  under the following assumption. The objective is the motion design  given the initial pose and optimal pose $\bm{\theta}_{op}$ in dynamic environments for  MDAMS. 

\begin{assumption}
	The global information of the dynamic environment is known in real time, i.e. the instantaneous positions and geometry characteristics of all obstacles are known, but the obstacles' future movements are unknown. 
\end{assumption}

\subsection{On-line motion planning algorithm}\label{sec:online}
\subsubsection{Overview}
\noindent
The \textbf{Algorithm \ref{alg:online}} conducts  motion planning and  execution  simultaneously by introducing three cycles: on-line sensing, collision-test and control cycles. 
In each sensing cycle, changes of the environment are captured and updated. The planning process re-plans the motion according to collision-test's output. In each control cycle, the robot switches to the above re-planned motion if there are predicted collisions.
\begin{algorithm}[t]
	\caption{On-line Motion Planning}\label{alg:online}
	\begin{algorithmic}[1]
		\STATE \emph{Initialization} : \emph{sensing} cycle $\Delta t_s $, 
		\emph{control} cycle  $\Delta t_c$,		 
		\emph{collision-test} cycle $\Delta t_{col} $, 
		$ iS  \leftarrow  1 $, $iC  \leftarrow  1$ 
		\STATE $ PosObsInit = EnvInit() $ \COMMENT{initialize the envitonment}
		\\ motion $ \leftarrow $ \textbf{Algorithm \ref{alg:offline}} \COMMENT{off-line motion planning}
		\WHILE{Goal not reached}
		\STATE \textbf{sensing:}
		\IF{start of the  \emph{sensing} cycle iS}  
		\STATE {$ PosObsNew = EnvUpdate() $} \COMMENT{update the environment information}
		\ELSIF{end of the  \emph{sensing} cycle iS}
		\STATE {iS $ \leftarrow $ iS+1}
		\ENDIF	
		\\ \textbf{planning:}
		\IF{start of the  \emph{control} cycle iC}  
		\STATE $ VelObsNew=ObsEstim(PosObsNew) $ \COMMENT{evaluate obstacles' movements}
		\STATE {$ CollisionCheck(PosObsNew,VelObsNew) $} \COMMENT{collision prediction}			   	  
		\IF{collision is predicted} 
		\STATE {$ motion  \leftarrow  On$-$linePlanning(\textbf{Algorithm \ref{alg:offline}}) $} \COMMENT{motion re-planning}
		\IF{a sudden collision appears}  
		\STATE {Immediate Stop} 
		\STATE  $ motion  \leftarrow UpdateMotion() $ \COMMENT{motion updating}
		\ENDIF
		\ENDIF
		\ELSIF{end of the  \emph{control} cycle iC}
		\STATE {iC $ \leftarrow $ iC+1}
		\IF{collision is predicted}
		\STATE $ motion  \leftarrow  UpdateMotion() $ \COMMENT{motion updating}
		\ENDIF
		\ENDIF
		\\ \textbf{tracking:}
		\STATE  $ MoveOn(motion) $ \COMMENT{motion tracking}		
		\ENDWHILE
	\end{algorithmic}
\end{algorithm}

\textbf{Algorithm \ref{alg:online}} starts by the initialization: the environmental information including positions of all obstacles is captured ($ EnvInit $). An off-line  motion is  designed using \textbf{Algorithm \ref{alg:offline}}.  The \emph{sensing}, \emph{control} and \emph{collision-test} cycles ( $\Delta t_{s} \leq \Delta t_c \leq \Delta t_{col} $) and  cycle numerations $ iS $ and $iC$ are initialized. Then,  MDAMS starts to track the designed motion and the on-line motion planning process is activated (\textbf{while} loop, lines $ 3 $ through $ 27 $). 

At the start of each \emph{sensing} cycle $t_{iS}$,  positions of all dynamic obstacles  are captured  (line $ 6 $, $EnvUpdate$) while conserving the historical ones at the same time. 

At the start of each \emph{control} cycle $t_{iC}$,  dynamic obstacles' future positions  are predicted  (line $ 11 $, $ObsEstim$, see Eqs. (\ref{eq:velobs}) and (\ref{eq:posobs})), and the obstacle-collision check is conducted for the current motion (line $ 12 $, $CollisionCheck$). 
If there are collisions predicted in the  \emph{collision-test} cycle $[t_{iC+1}, t_{iC+1} +\Delta t_{col} ]$, the on-line motion re-planning process  $On$-$linePlanning$ will be called (line $ 14 $). 

It is worth mentioning that the start-node of $On$-$linePlanning$ is  $ \bm{q}(t_{iC+1}) $ of the next \emph{control} cycle $[t_{iC+1}, t_{iC+2}]$. 
The start time $ t_{start} $ of   $On$-$linePlanning$ is  $t_{iC+1}$ of the next \emph{control} cycle $[t_{iC+1}, t_{iC+2}]$.
In particular,  if a sudden collision appears (line $ 15 $), the MDAMS will stop (line $ 16 $) immediately and update the motion  (line $ 17 $). In that case, the start-node  of  $On$-$linePlanning$ will be the current state $ \bm{q}(t) $, and the re-planning start time $ t_{start} $ will be the current time $t$. 

At the end of each \emph{control} cycle $ t_{iC+1} $, the motion will be updated (line $ 23 $) if  collisions are predicted (line $ 12 $). As time goes on, the motion tracking is done continuously (line $ 26 $, $MoveOn$).

\subsubsection{Obstacle's motion prediction}
\noindent
In \textbf{Assumption 1}, we suppose that dynamic obstacles' positions  are known in real time. The  dynamic obstacles' motions are estimated  based on their historical positions (line $ 11 $, $ObsEstim$ in \textbf{Algorithm \ref{alg:online}}). Specifically, the velocity of dynamic obstacle $ i $  is estimated  as follows:
\begin{equation}\label{eq:velobs}
\bm{V}_{oi} = max \{\|({\bm{X}_{oi}(t_{k})-\bm{X}_{oi}(t_{k-1})})/{\Delta t_{s}}\|, \; k = iS-m_o,...,iS\}
\end{equation}
where $m_o$ is the numeration  of historical positions of obstacle $ i $, $ \bm{X}_{oi}(t_{i}) $ is the position of obstacle $i$  at time $t_{i}$, $ \Delta t_{s} $ is the \emph{sensing} cycle and  $ iS $ is the \emph{sensing} cycle numeration.
Then, the position of obstacle $ i $ at time $ t= t_{iC}+\delta t $ can be estimated (line $ 12 $, $CollisionCheck$) as  
\begin{equation}\label{eq:posobs}
\bm{X}_{oi}(t_{iC}+\delta t) = \bm{X}_{oi}(t_{iC}) + \delta t \bm{V}_{oi}
\end{equation}
where $ iC $ is the \emph{control} cycle numeration,  $\delta t\in (0, \Delta t_{col}]$. 

Recall that in \textbf{Assumption 1}, the global environment information is assumed to be known  in real time, which is sometimes difficult to obtain.  
In that case, the proposed on-line motion planning \textbf{Algorithm \ref{alg:online}} will lose its efficiency. 	With regard to this, the following relaxed assumption is presented.
\begin{assumption}
	Only the local environment information within the sensing range is known. 
\end{assumption}	
The  \textbf{Algorithm \ref{alg:online}} is modified as follows. The environment out of  sensing range is treated as a collision-free space. Then, a sequence of via-points $\{\bm{q}_i, i = 1,...,n_p\}$  bypassing the sensed obstacles can be found using \textbf{Algorithm \ref{alg:offline}} (e.g., four via-points \{$ \bm{q}_{init} $, $ \bm{D} $, $ \bm{C} $, $ \bm{q}_{goal} $\} in Fig. \ref{fig:MBOptim}). At the beginning, the robot tries to move along the first segment $ \bm{q}_{init} $-$ \bm{D} $.  Then, the environment will be updated as time goes on. This process will repeat until the robot arrives at $ \bm{q}_{goal} $.  

\subsection{EEs' via-point-based MOGA motion planning}\label{sec:via}
\noindent 
Though the proposed \textbf{Algorithm \ref{alg:online}}  can be used to design the motion for any DoF mechanical systems, the collision-free test  for a high DoF system is time-consuming and complex. What is worse,  the direct motion planning in joint space will lead to unforeseen behaviors in task space, and the via-poses are not considered.
Regarding to this, the objective of this subsection is to propose an EEs' via-point-based MOGA motion planning algorithm to optimize the via-poses given EEs' desired positions-orientations.

\subsubsection{Algorithm}
\noindent
The mobile base's trajectory $\bm{\theta}_m = (x,y,\phi)$ and the joint trajectory of manipulator $(\bm{\theta}_w, \bm{\theta}_R, \bm{\theta}_L)$ characterize the MDAMS's motion  (see Section \ref{sec:2}). Recall that the initial pose and EEs' desired positions-orientations are known in~Section \ref{sec:2}. A sequence of collision-free via-points $\{\bm{X}_i=[\bm{X}_{R_i}^T\, \bm{X}_{L_i}^T]^T \in \mathbb{R}^{4m\times 1},\, i = 1,...,n_p\}$ are designed  using   \textbf{Algorithm \ref{alg:online}}, where $ \bm{X}_{R_i} $ and $ \bm{X}_{L_i} $ are EEs' via-points, and $n_p$ is the number of  EEs' via-points. 
The objective is to design optimal via-poses $\bm{\Theta}_i=(x_i,y_i,\phi_i,\bm{\theta}_{w_i}, \bm{\theta}_{R_i}, \bm{\theta}_{L_i})$ corresponding to $\{\bm{X}_i\}$. 

MOGAs (e.g. the improved MaxiMin NSGA-II  \textbf{algorithm \ref{alg:MaxiMinNSGA}}) are used to search for the optimal via-poses.
The chromosome  is chosen as $\{\bm{\Theta}_{1},\,...\,,\bm{\Theta}_{n_p}\}$. The searching interval for each via-pose  is $  \{\bm{\Theta}_{ij} \in[\bm{\theta}_{jmin}, \bm{\theta}_{jmax}], j = 1,...,n\} $, where $\bm{\theta}_{j_{max}}$ and $ \bm{\theta}_{j_{min}}$ are respectively the upper and lower boundaries of  $ \bm{\theta}_j $ (see Section \ref{sec:2}). 
Specifically,   $(x_i,y_i)$  is searched within the reaching region of  via-point $\{\bm{X}_i\}$,  and $\phi_i$  is searched within $[-\pi, \pi]$.
The via-pose-based objective functions are defined in the next subsection.

\subsubsection{Via-pose-based objective functions} \label{sec:moga}
\begin{itemize}
	\item \textbf{EE's positioning accuracy}:
	The via-poses $\{\bm{\Theta}_i\}$ are required to bypass all the EE's via-points $ \{\bm{X}_i\} $. Hence, the first objective function  is defined as follows:\\
	$f_a(\bm{\Theta}) = \sum_{i=1}^{n_p}\|{\bm{X}_i - \bm{X}_i(\bm{\Theta}_{i})}\|$,    
	where $\bm{X}_i(\bm{\Theta}_{i})$ is the  EEs' via-points corresponding to  via-pose $ \bm{\Theta}_{i} $  based on the forward kinematics expressed by $ T_{R}$ and $ T_{L} $ in Section \ref{sec:2}.
	\item \textbf{Joint displacement}: The least joint displacement is required to minimize energy consumption. Then, \\
	$f_b(\bm{\Theta}) = \sum_{i=1}^{n_p} \sum_{j=1}^{n} \|({\bm{\Theta}_{ij} - \bm{\Theta}_{(i-1)j}})/({\bm{\theta}_{j_{max}} - \bm{\theta}_{j_{min}}})\|$. 
	\item \textbf{Collision evaluation}:
	In order to take into account the robot-obstacle intersection,
	simplify the MDAMS as a skeleton in Fig. \ref{fig:robotmodel} (b) and introduce  eight control points ($ 1,...,8 $) and eight links ($a,...,h$), 
	i.e. the CoMs of mobile base, waist, two shoulders, two elbows and two wrists, and  eight connection links between them.
	Then, define the third objective function as follows:\\
	$f_c(\bm{\Theta}) = \sum_{i=1}^{n_p}\sum_{j=1}^{n_c}{Lc_{ij}(\bm{\Theta_i}) }/{L}$, where $ n_c $ is the number of links on MDAMS, $Lc_{ij}(\bm{\Theta_i})$ is the  length of  link which collides with obstacles along the link $j$, and $ {L} $ is the length of eight defined links. Take the situation in  Fig. \ref{fig:robotmodel} (b) as an example: the left forearm intersects an obstacle, then the third objective function is given as $ f_c = {h}/({a+b+ \cdot\cdot \cdot +h}) $ for one via-pose, $ n_p = 1$.
	\item \textbf{Directional manipulability}:
	To facilitate the motion  bypassing all the EEs' via-points $ \{\bm{X}_{i}\} $, there is a directional manipulability preference between two adjacent via-points $ {\bm{X}}_{i}  $ and $ {\bm{X}}_{i+1}  $  in task space. Then,\\
	$f_d(\bm{\Theta}) = -\sum_{i=1}^{n_p-1}(\Omega_{Bdir}(\bm{d}_{R_i})+\Omega_{Bdir}(\bm{d}_{L_i}))$, where $\Omega_{Bdir}(\bm{d}_{R_i})$ and $ \Omega_{Bdir}(\bm{d}_{L_i}) $  are respectively the directional manipulabilities of right and left EEs which are defined as follows: 
	\begin{equation}
	\Omega_{Bdir}({\bm{d}_{\kappa_i}})=\sum_{k=1}^{m} \|(\bm{d}_{\kappa_i}^T \cdot \bm{u}_{ik}) \sigma_{ik}\|
	\end{equation}
	where $\bm{d}_{\kappa_i}= ({{{x}}_{\kappa}({\bm\Theta}_{i+1}) - {{x}}_{\kappa}(\bm{\Theta}_{i}) })/{\|{{x}}_{\kappa}(\bm{\Theta}_{i+1}) - {{x}}_{\kappa}(\bm{\Theta}_{i}) \|} $ is the unit vector  along each segment $\{{\bm{X}}_{i}(\bm{\Theta}_{i})-\bm{X}_{i+1}(\bm{\Theta}_{i+1}), i = 1,...,n_p-1 \}$.  $ (\cdot) $ is the dot product and $ x_{\kappa}, \kappa\in \{R, L\}$ is defined in~Section \ref{sec:2}. 
	Decompose Jacobian $J_{\kappa}(\bm{\Theta}_{i})\in \mathbb{R}^{m \times n}, \kappa\in \{R, L\} $ in~Section \ref{sec:2} using the singular value decomposition technique $J_{\kappa}(\bm{\Theta}_{i}) = U_i\Sigma_i V_i$,  $\bm{u}_{ik}$ is the $k$-th column vector of  $ U_i \in \mathbb{R}^{m \times m}$, $ \sigma_{ik} $ is the $k$-th singular value of  $\Sigma_i \in \mathbb{R}^{m \times n}$ and  $V_i \in \mathbb{R}^{n \times n}$.  
\end{itemize}
\noindent
As a result, the optimization problem is formulated as follows:
\begin{equation}
\begin{array}{c}
min (f_a(\bm{\Theta}), f_b(\bm{\Theta}), f_c(\bm{\Theta}),f_d(\bm{\Theta}))\\
s.t. \; \bm{\Theta} \in C_{free}
\end{array}.
\end{equation}

\section{Simulations Results}\label{sec:5}

\subsection{Optimal pose design and path planning}
\noindent
In Fig. \ref{fig:motAd}, a table is surrounded by four chairs $ A $, $ B $, $ C $ and $ D $. 
The  task is to design the motion for MDAMS to manipulate chair $A$, which can be divided into two subtasks: the \emph{optimal pose} $\bm{\theta}_{op}$ design around chair $A$ and  the collision-free approaching motion design.
The EEs' desired positions are known as  $ x_{Rd} = (2.7,\, 0.3,\,0)$, $x_{Ld} = (2.7,\,-0.3,\,0) $, and the initial positions are  $ x_{R} = (-0.2,\, 0.25,\, -0.219)$, $x_{L} = (-0.2,\, 0.55,\,-0.219) $. 
Based on the robot's dimension in Table $ 1 $, the expected position and orientation of mobile base are calculated as $(x_{me}, y_{me}) =(3,\,   0) $ and  $ \phi_{de} = 180^{\circ}  $, respectively.
Set the initial pose of MDAMS as $\bm{\theta}_{init} =(-0.2, 0.4,0,\bm{0}_{16})$.


\subsubsection{{Optimal pose} design}\label{sec:chairA}
\begin{figure}[t]
	\centerline{\psfig{file=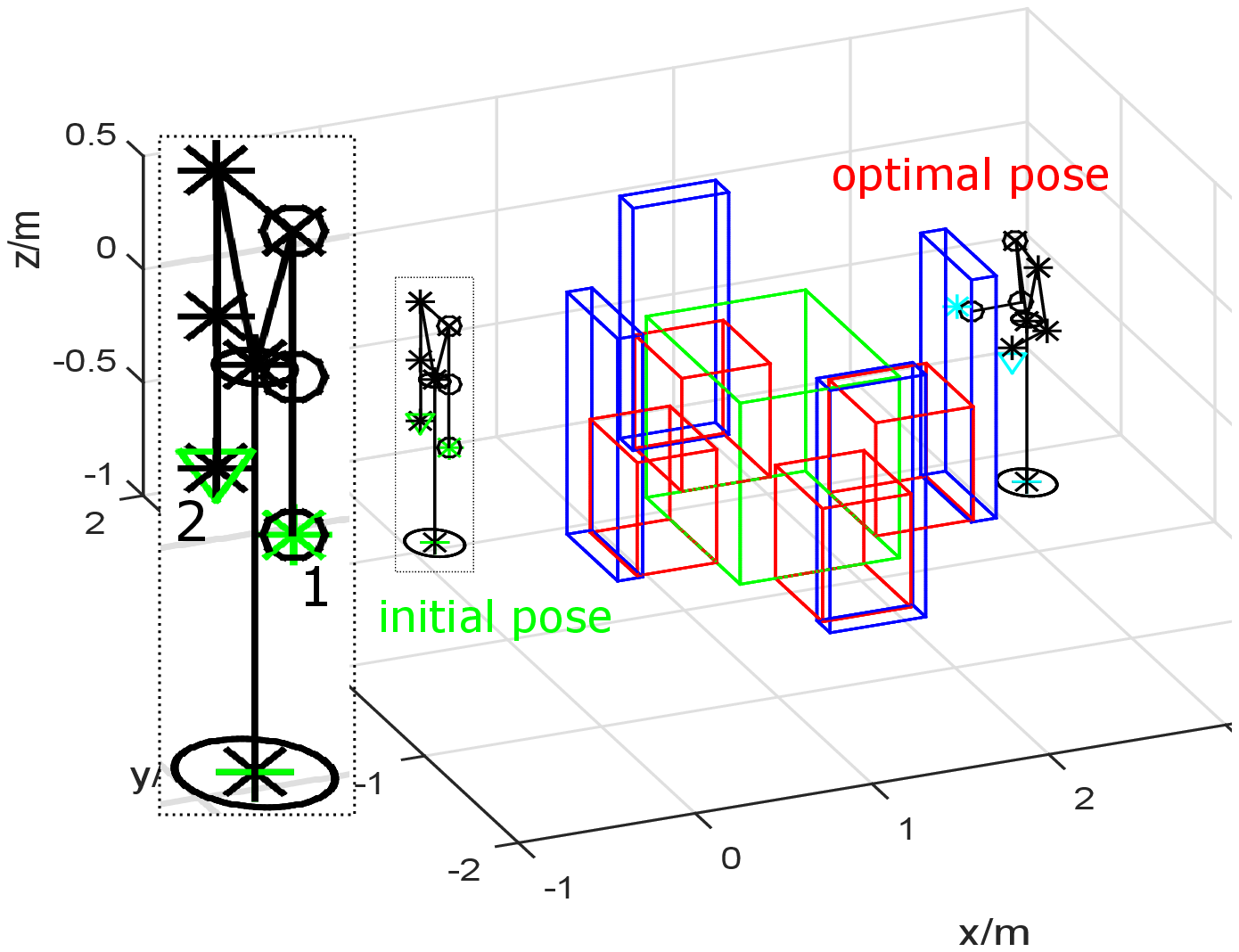,width=6.5cm}\psfig{file=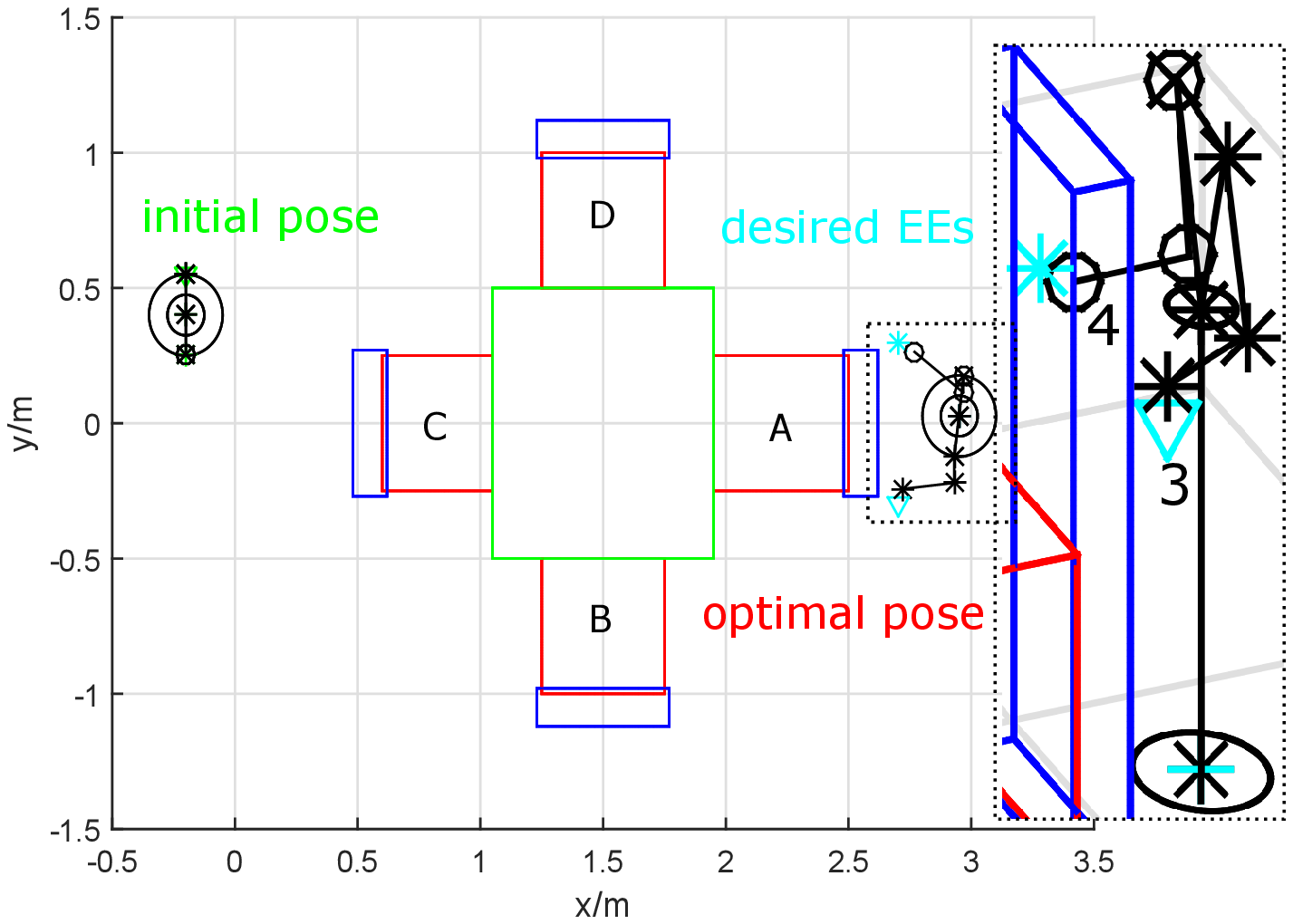,width=6cm}} 
	\centerline{\psfig{file=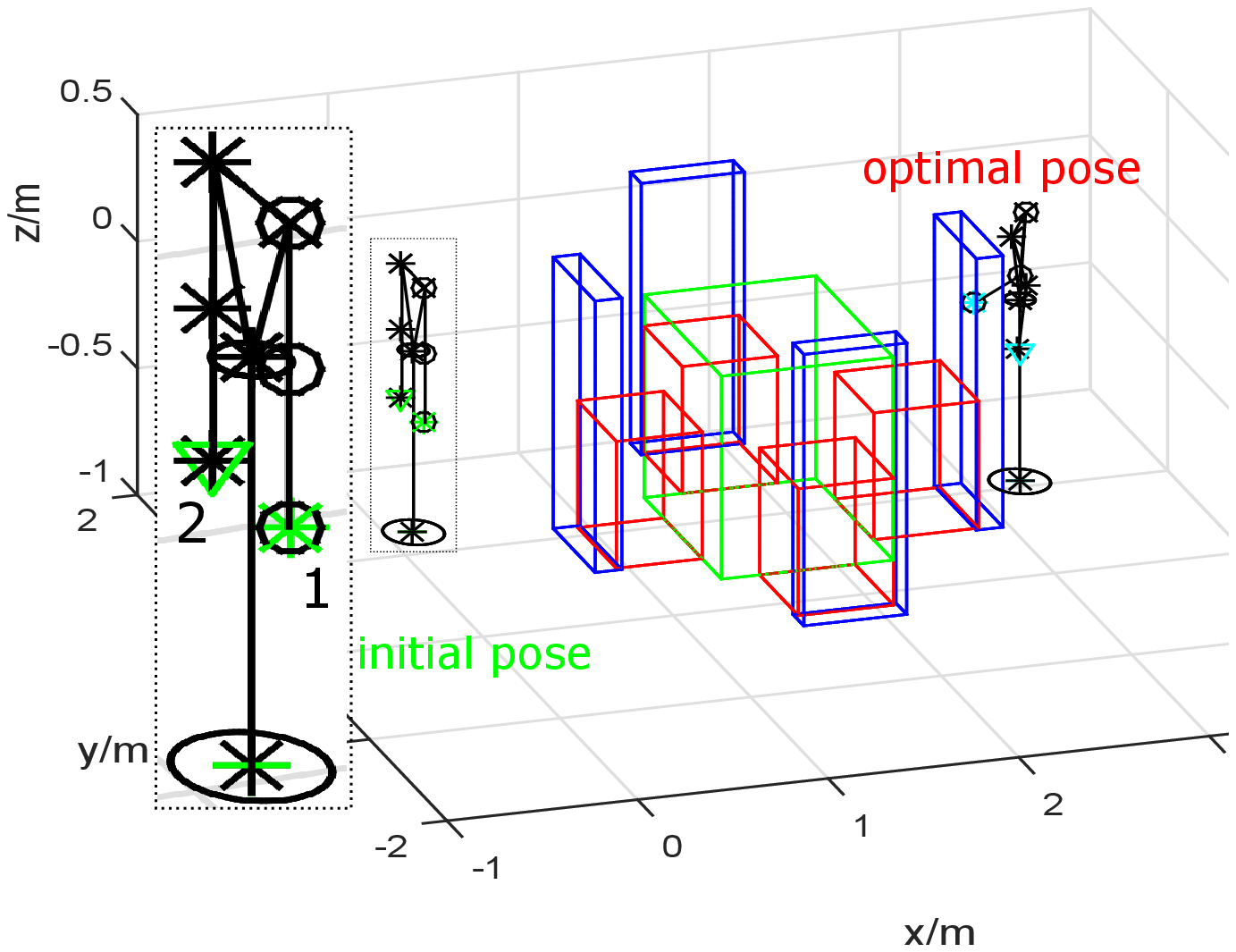,width=6.5cm}\psfig{file=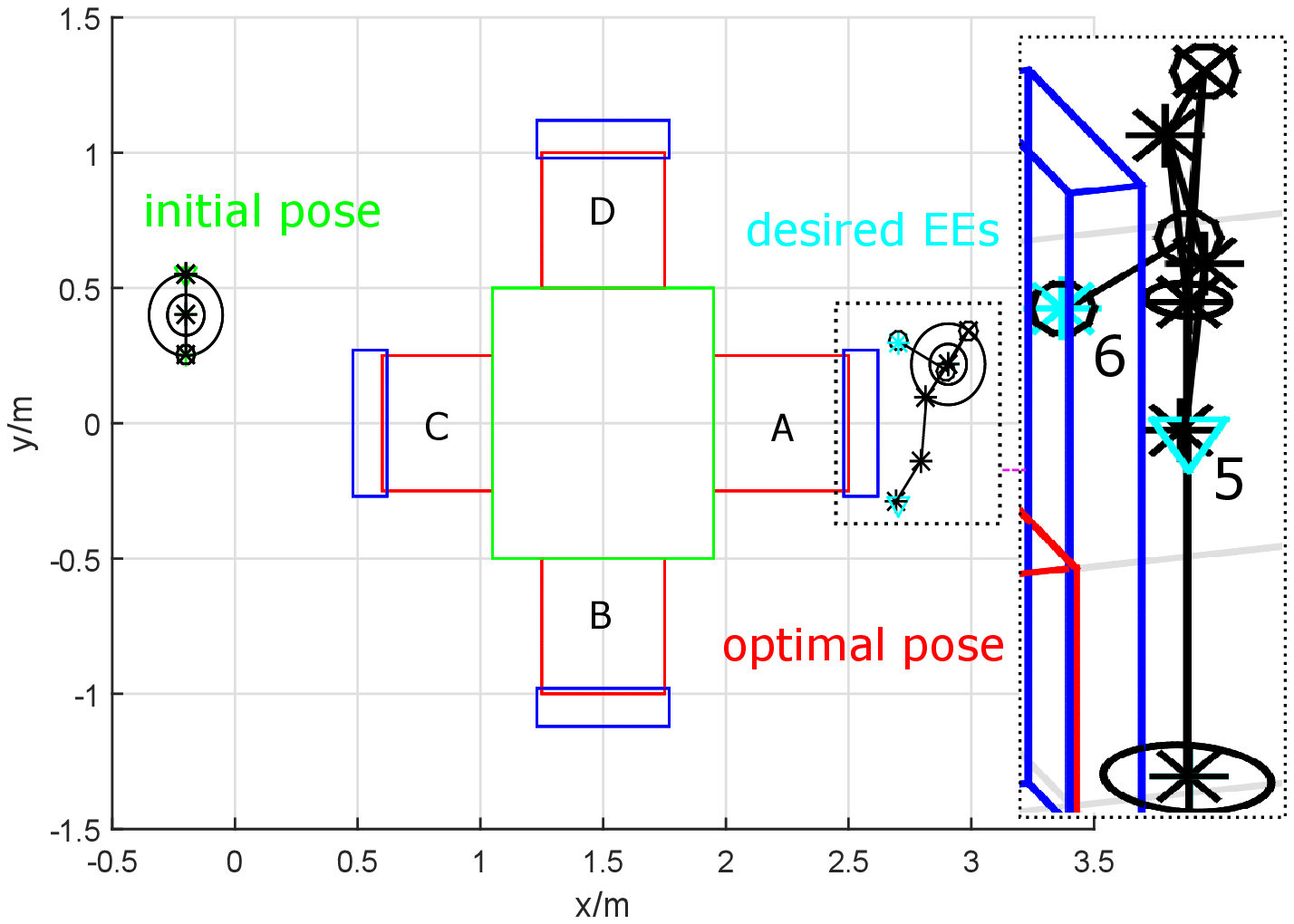,width=6cm}} 
	\vspace*{8pt}
	\caption{Pose design of MDAMS to manipulate chair $A$. Up: improved MaxiMin NSGA-II algorithm. Down: combined fitness function-based MOGA.}
	\label{fig:motAd}
\end{figure}

\noindent
This subsection
validates the \textbf{Algorithm \ref{alg:MaxiMinNSGA}}.
Set the  population size  as $N_{pop}=100$,  the generation number as $N_G = 120$ and the genetic crossover and mutation probabilities as $p_c = 0.7$ and $p_m = 0.3$.  According to the importance of each objective, the  coefficients in Eq. (\ref{eq:fitdis}) are chosen as $w_i=(0.54,\,0,\,0.04,\,0.02,\,0.4), i=1,\ldots, N_{pop}$.
After $N_G$ generations, the designed \emph{optimal pose} $\bm{\theta}_{op}$ is shown in  Table $ 2 $  and Fig. \ref{fig:motAd}.

\begin{table}[t]
	\centering
	\caption{The designed optimal pose $\bm{\theta}_{op}$ $ [m, deg] $.}
	\label{tbl:gaA1}
	\begin{tabular}{c|r|r|r|r|r|r|r}
		\hline
		$\bm{\theta}_{m}$ \& $\bm{\theta}_{w}$ & 2.95  &  0.027 &      166.77 &        5.90  &  0.24 & -& -\\ \hline
		
		$\bm{\theta}_{R}$ & 1.11 &  13.73  & 33.19 &  74.63   & 3.85 &  49.38 &  {4.81} \\ \hline
		$\bm{\theta}_{L}$ & 2.77  & 22.07& 8.19 &  64.46 &  67.65 &  67.63 &  {23.91} \\ \hline		
	\end{tabular} 
\end{table}

It can be seen that the designed optimal position of mobile base is $ (x_{md},y_{md}) = (2.95,\,   0.027) $ which is near the expected position $ (x_{me}, y_{me}) $, the positioning error is $(-0.05,\,   0.027)$; the designed optimal orientation of  mobile base is $ \phi_d = 166.77 ^{\circ} $ which is near the expected orientation $ \phi_{de} $, the orientation error is $  0.23 \, rad$.

The up two figures in Fig. \ref{fig:motAd}  show the motion planning results using the proposed improved MaxiMin NSGA-II \textbf{Algorithm \ref{alg:MaxiMinNSGA}}.
The black skeletons represent the designed MDAMS. Points $1$ and $2$  show  the initial positions of  right and left EEs. 
The cyan plus shows  the designed optimal position $ (x_{md},y_{md}) $ of mobile base.  
Points $3$ and $4$  show the designed  positions $( x_{R}(\bm{\theta}_{op}), x_{L}(\bm{\theta}_{op}) )$ of right and left EEs which can be calculated according to the forward kinematics.
It can be seen that the designed optimal pose $\bm{\theta}_{op}$ is ``human-like'' with good positioning accuracy to EEs' desired positions $( x_{Rd}, x_{Ld} )$ in the objective function $ f_1 $ of Subsection~\ref{Sec:Objfunc}.
\begin{figure}[t]
	\centerline{\psfig{file=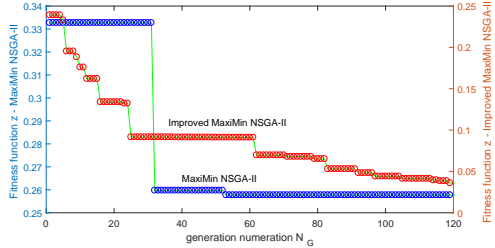,width=7cm}} 
	\vspace*{8pt}
	\caption{Comparison between the improved MaxiMin NSGA-II  and  MaxiMin NSGA-II  algorithms.}
	\label{fig:maximinCom}
\end{figure}
As a comparison,  the motion planning results using the combined fitness function-based MOGA  are shown in the down two figures in Fig. \ref{fig:motAd} with the combined fitness function being defined in Eq. (\ref{eq:fitdis}). It can be seen that even though the EEs' positioning accuracy is better than that of the \textbf{Algorithm \ref{alg:MaxiMinNSGA}}, the designed optimal mobile base's position-orientation and manipulator's configuration are twisted, not  ``human-like'' and far away from expectations.

Furthermore, two  optimal pose design simulations are realized in Fig. \ref{fig:maximinCom} which shows that the  optimal pose's fitness function value evolution of Eq. ~\eqref{eq:fitdis}. The optimal solution converges more quickly and continuously with better performance using the  \textbf{Algorithm \ref{alg:MaxiMinNSGA}}.

\subsubsection{Path planning and optimization for the mobile base}
\noindent
This subsection validates the path planning  \textbf{Algorithm \ref{alg:offline}}.
The mobile base has a circle geometry. To guarantee the collision-free navigation for the mobile base from the initial position to the designed optimal position $(x_{md},y_{md})$, we enlarge the obstacles by a security hull $\delta_{sm} = 0.3m$ including the mobile base's dimension. 
\begin{figure}[th]
	\centerline{{\scriptsize (a)}\psfig{file=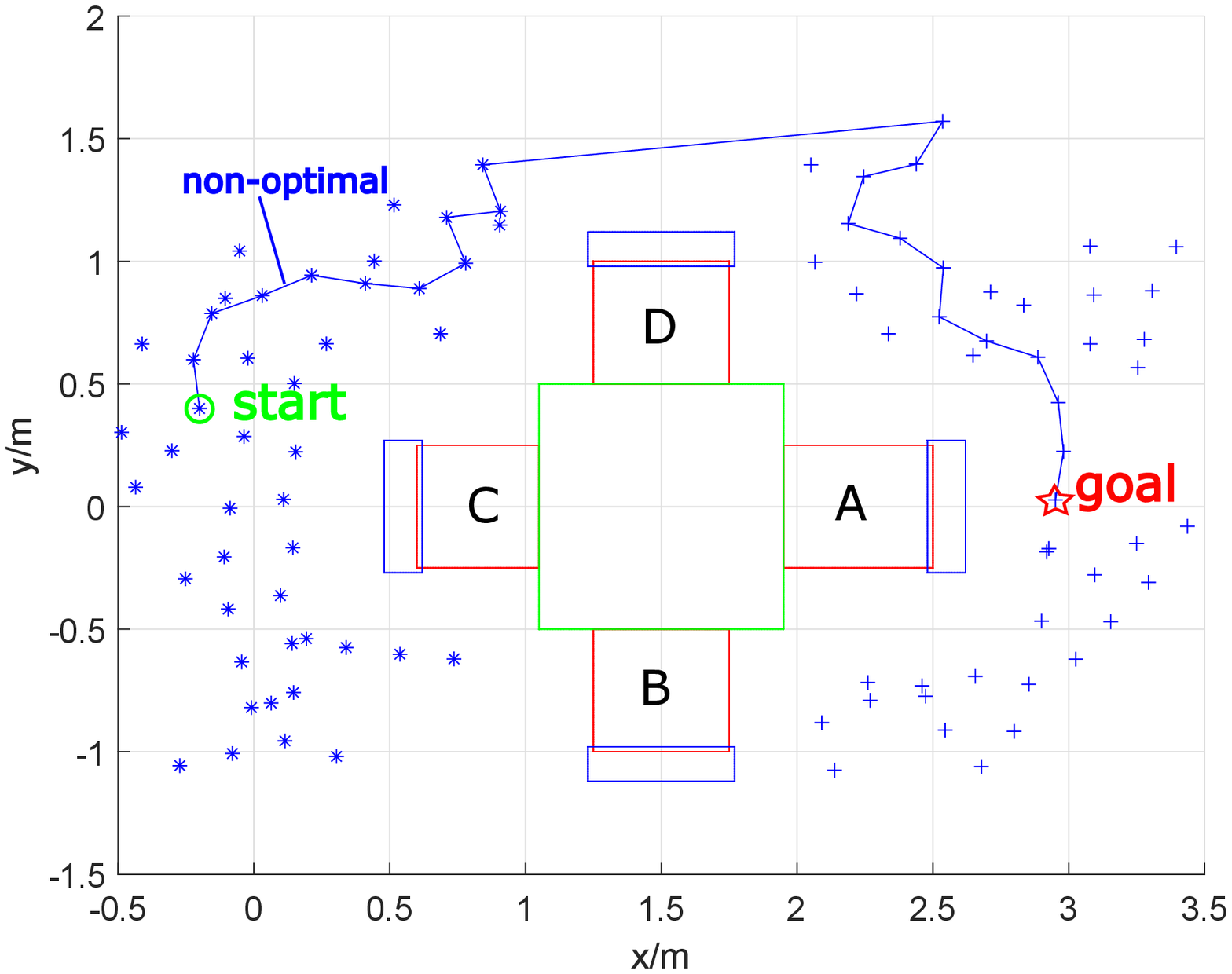,width=6cm}{\scriptsize (b)}\psfig{file=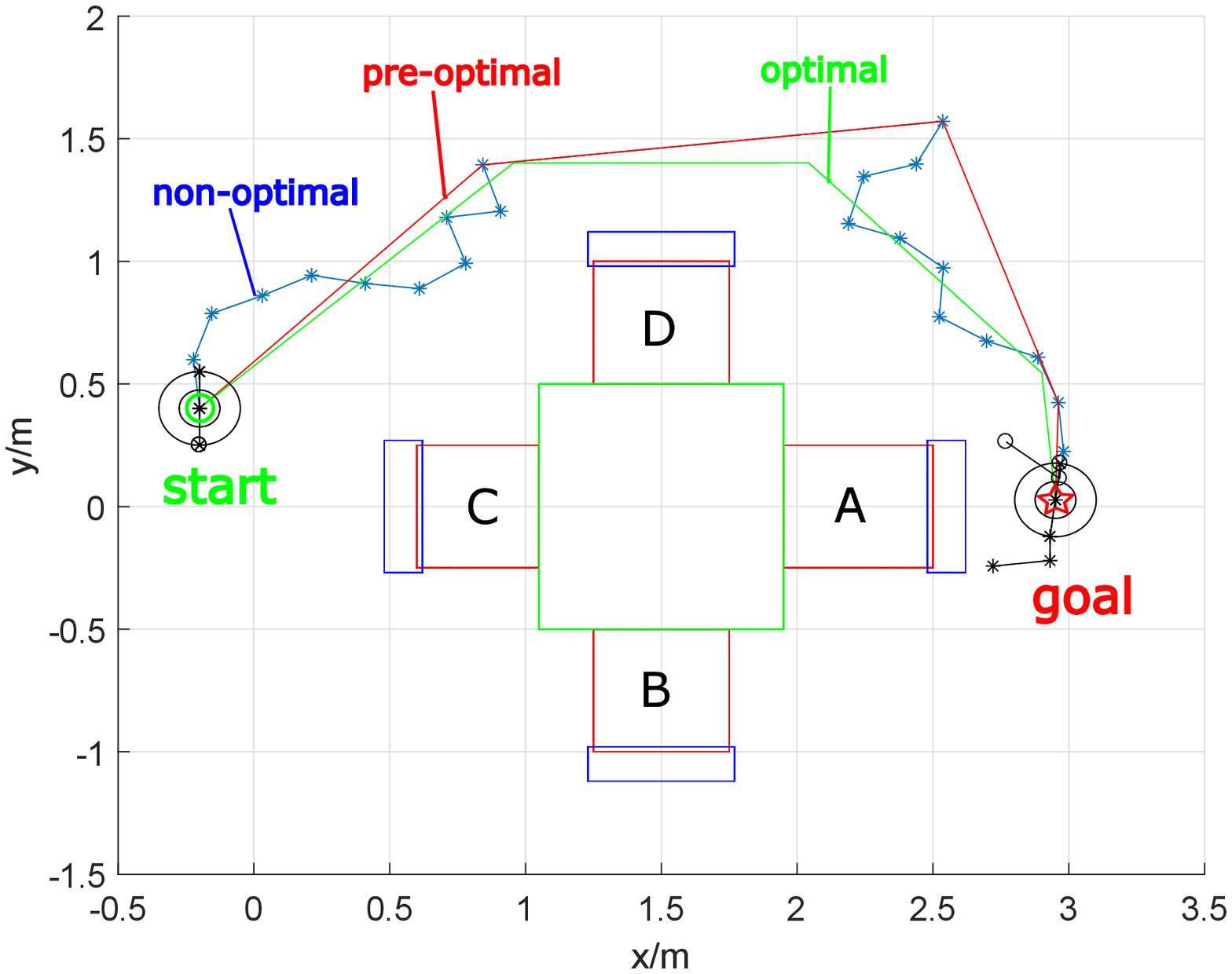,width=6cm}} 
	\vspace*{8pt}
	\caption{Path planning results for the mobile base: (a) without geometric optimization, (b) with geometric optimization.}
	\label{fig:tpA}
\end{figure}
The Fig. {\ref{fig:tpA}} (a) shows the designed path without geometric optimization. 
The blue star points, blue plus points and the blue line
represent the sampling nodes of the start exploring tree $ Tr_s $, the end exploring tree $ Tr_e $ and the designed path, respectively. It can be seen that two exploring trees are connected directly once there is a collision-free connection between them even if they are still very far away from each other. 
A collision-free approaching path is designed linking the mobile base's initial and desired positions, however, it is  not the optimal path.

The Fig. {\ref{fig:tpA}} (b) shows the  path optimization results using the proposed geometric optimization method. 
The red line represents the \emph{pre-optimal path} after \emph{node rejection}, the green line represents the \emph{optimal path} after \emph{node adjustment} which is also the designed optimal path. 
In the optimal path, there are in total $n_p=5$  via-points which can be adjusted within their neighborhoods to increase the clearance.

\subsection{On-line motion planning for the mobile base}
\noindent
This subsection is to verify the \textbf{Algorithm \ref{alg:online}}.  Note that motion tracking  of the mobile base is not considered in this paper.
The dynamic obstacles  move on the ground with bounded  stochastic velocities.
Set 
the \emph{sensing}, \emph{control} and \emph{collision-test} cycles as $\Delta t_s = 0.5s$, $\Delta t_c = 1s$ and $\Delta t_{col} = 2s$. 
In the following,  on-line motion planning for the mobile base is  realized  using Matlab in a standard PC (Intel(R) Core(TM) i5-3317U CPU @ 1.70GHz 1.70 GHz, 4,00Go, x64). 
\begin{figure}[th]
	\centerline{{\scriptsize (a)}\psfig{file=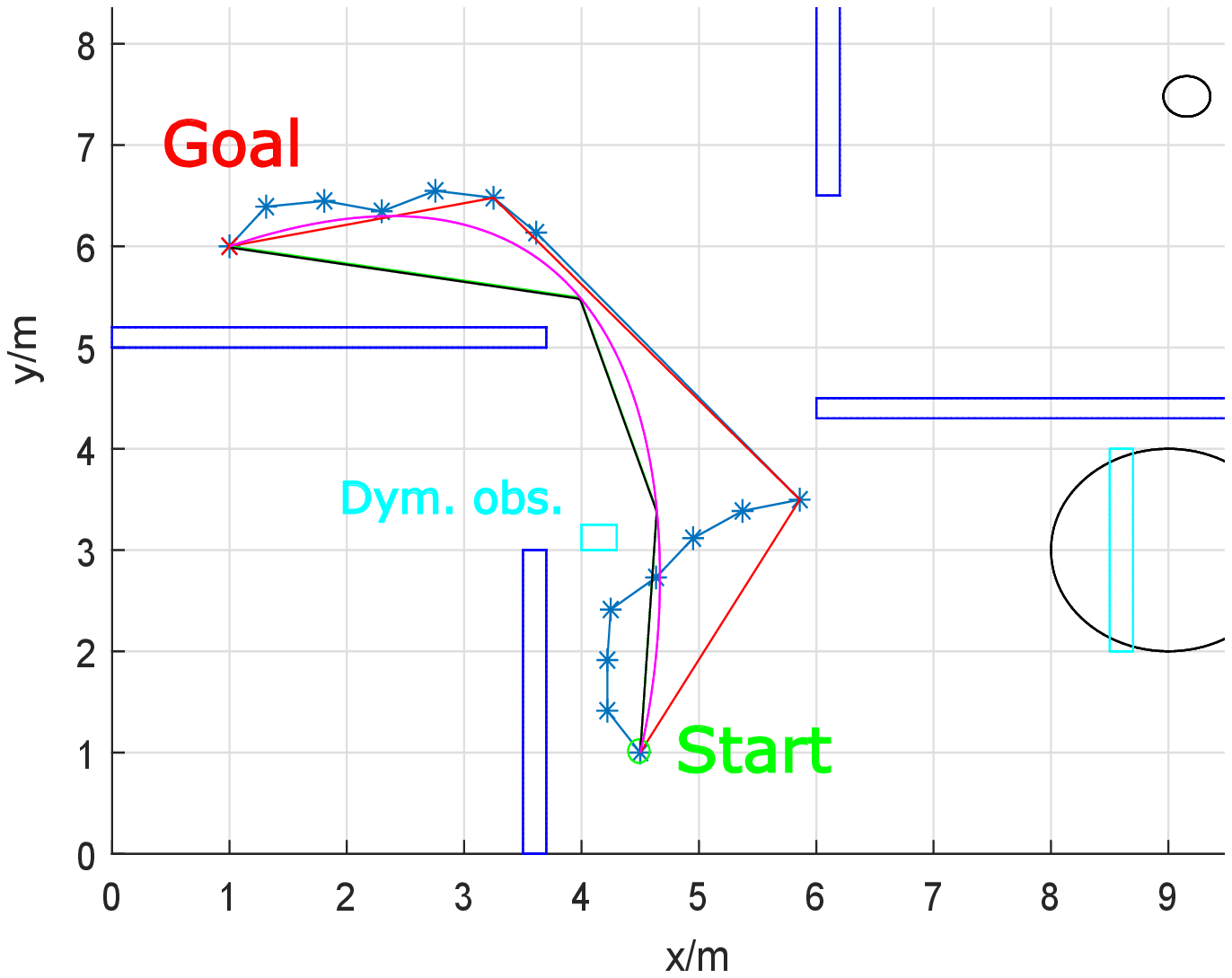,width=6cm}{\scriptsize (b)}\psfig{file=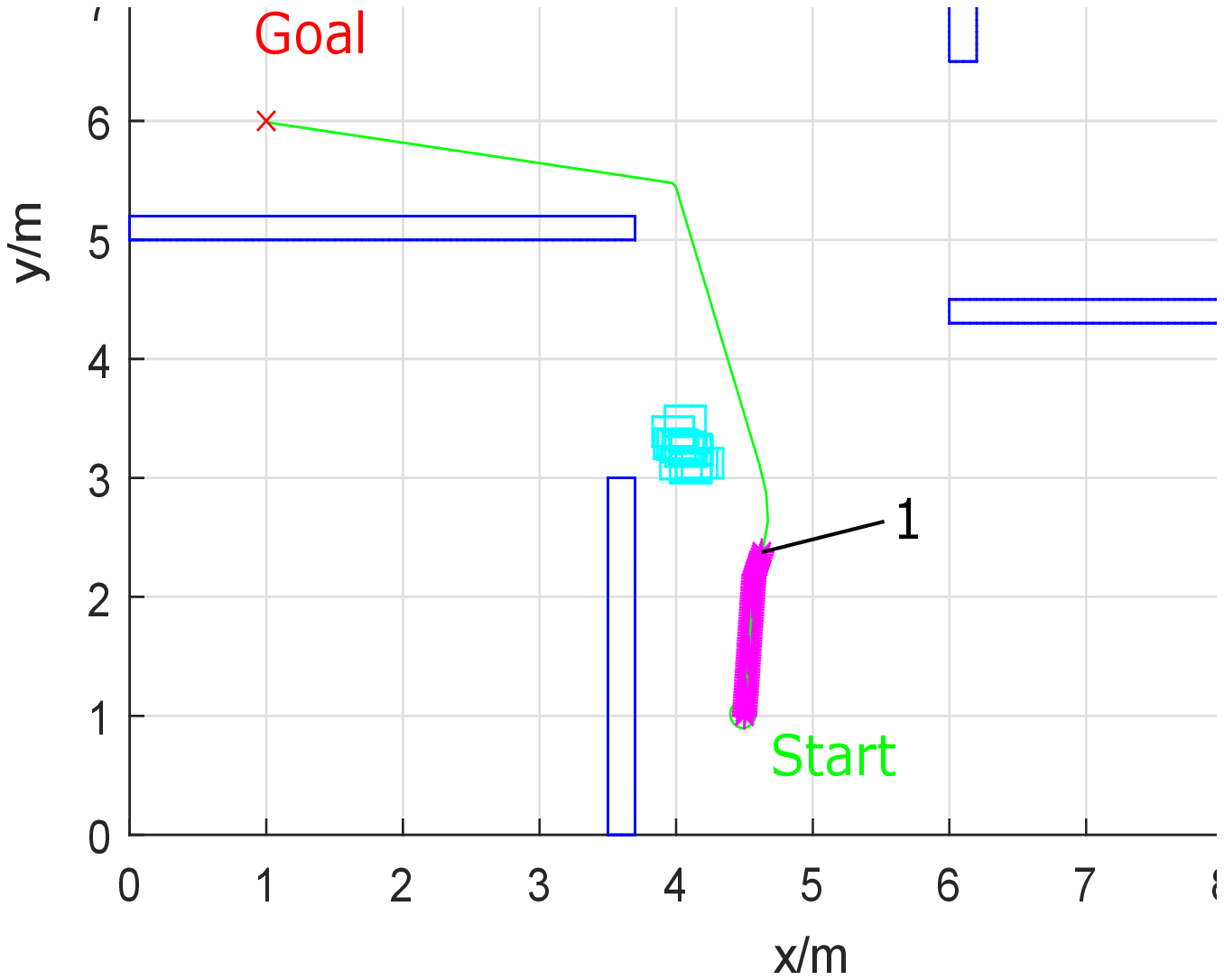,width=6cm}} 
	\centerline{{\scriptsize (c)}\psfig{file=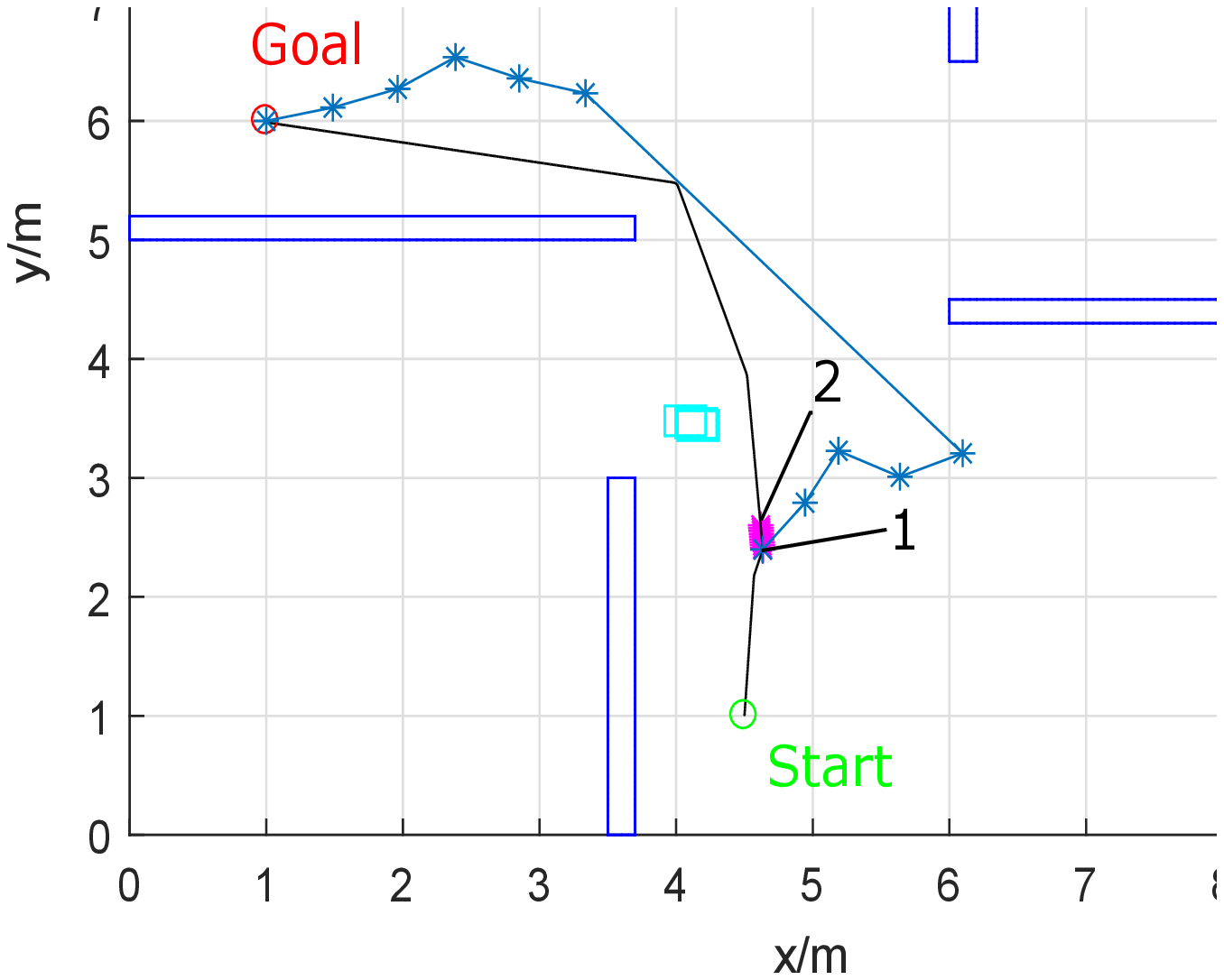,width=6cm}{\scriptsize (d)}\psfig{file=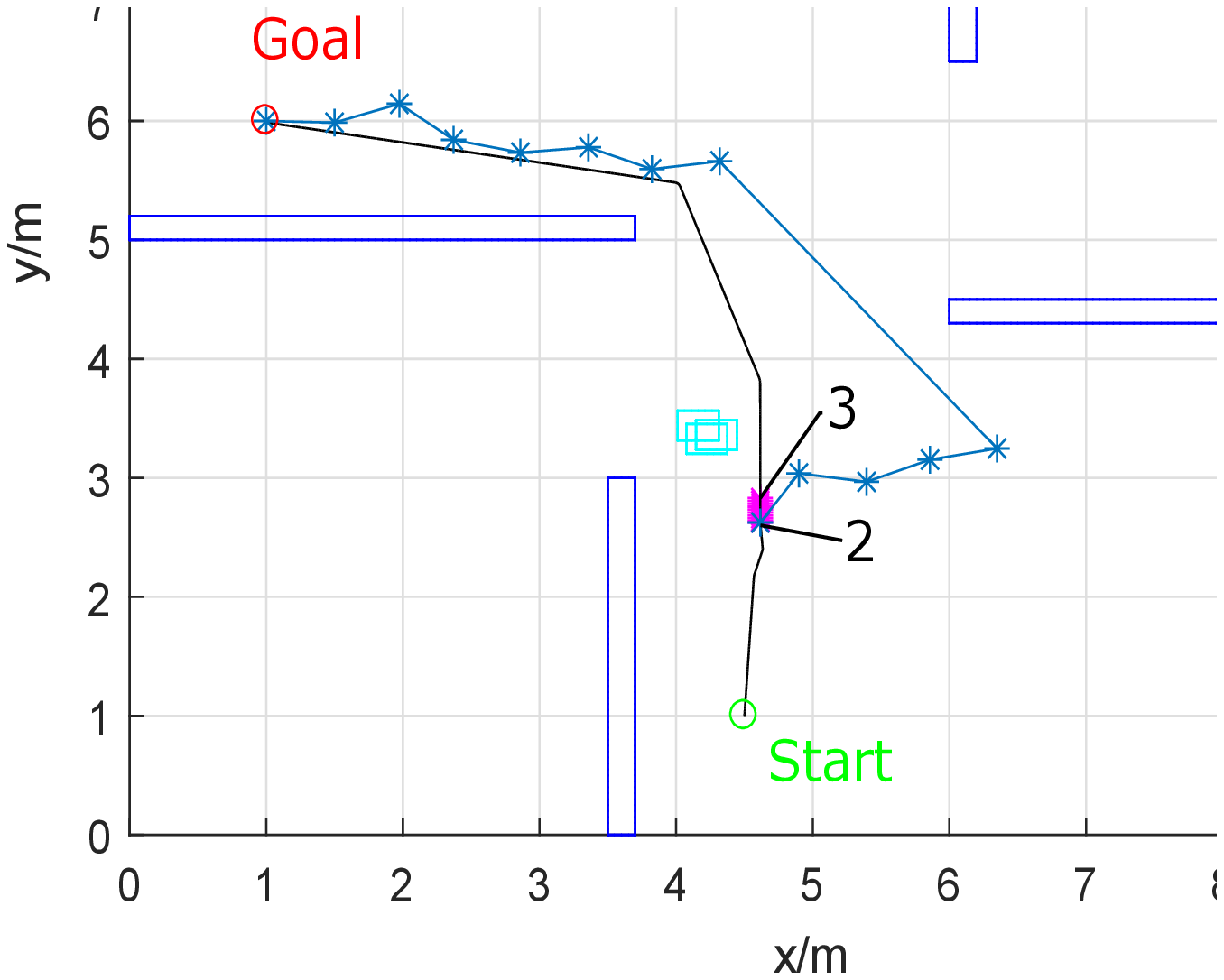,width=6cm}} 
	\caption{On-line motion planning results for  mobile base (I). (a) Off-line motion planning results, (b) Phase $ 1 $, (c) Phase $ 2 $, (d) Phase $ 3 $.}
	\label{fig:test2da}
\end{figure}
\begin{figure}[th]
	\centerline{{\scriptsize (a)}\psfig{file=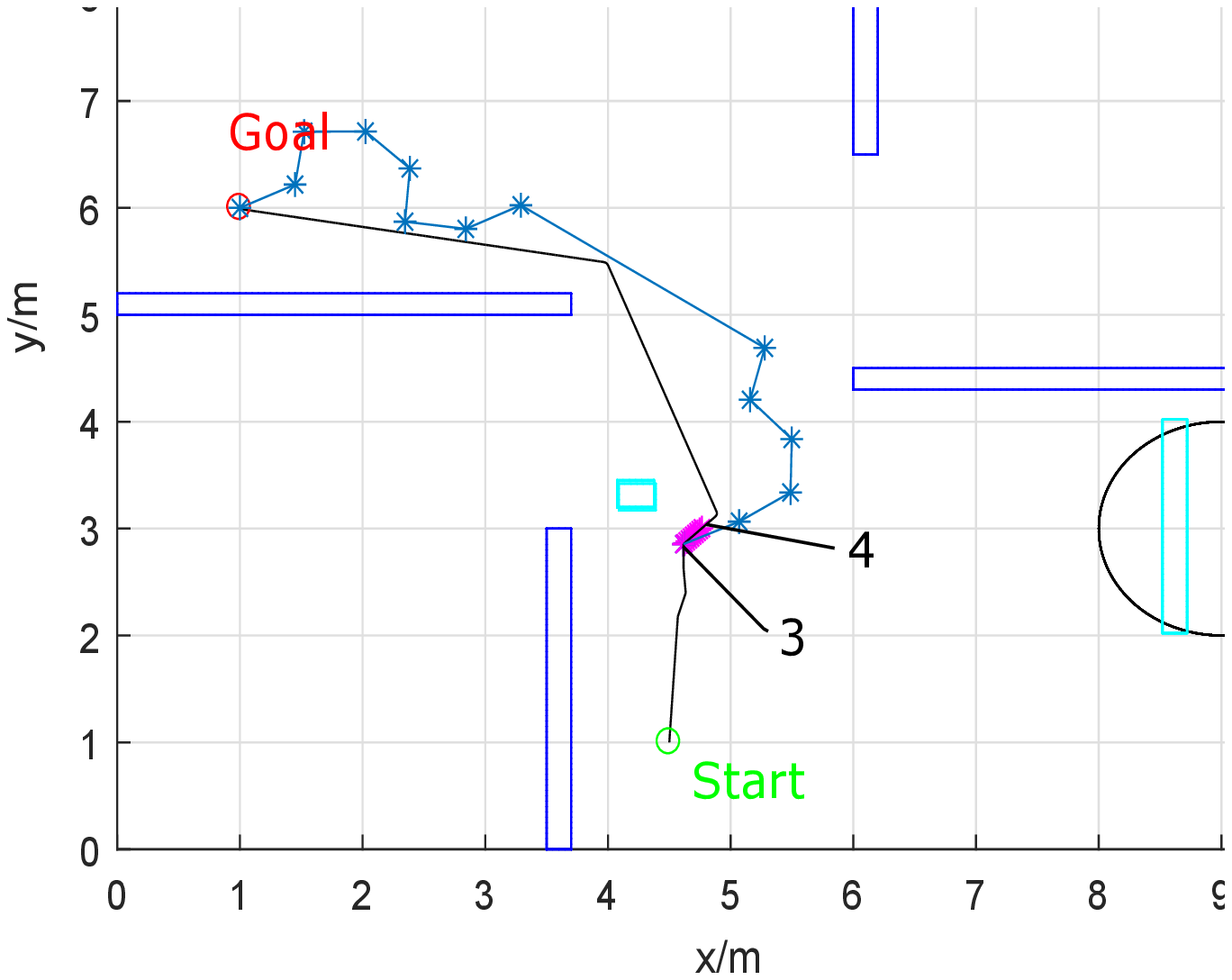,width=4cm}{\scriptsize (b)}\psfig{file=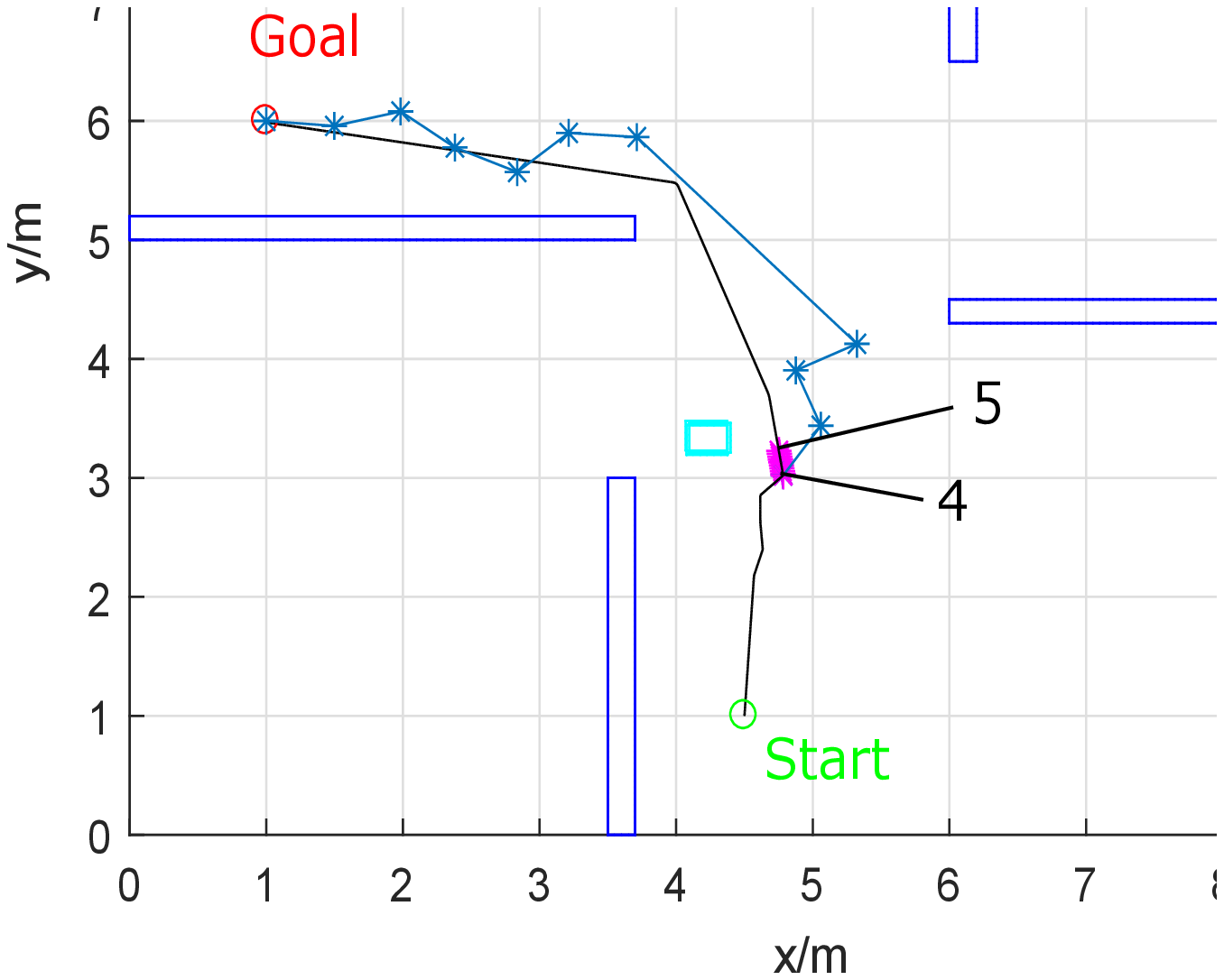,width=4cm}{\scriptsize (c)}\psfig{file=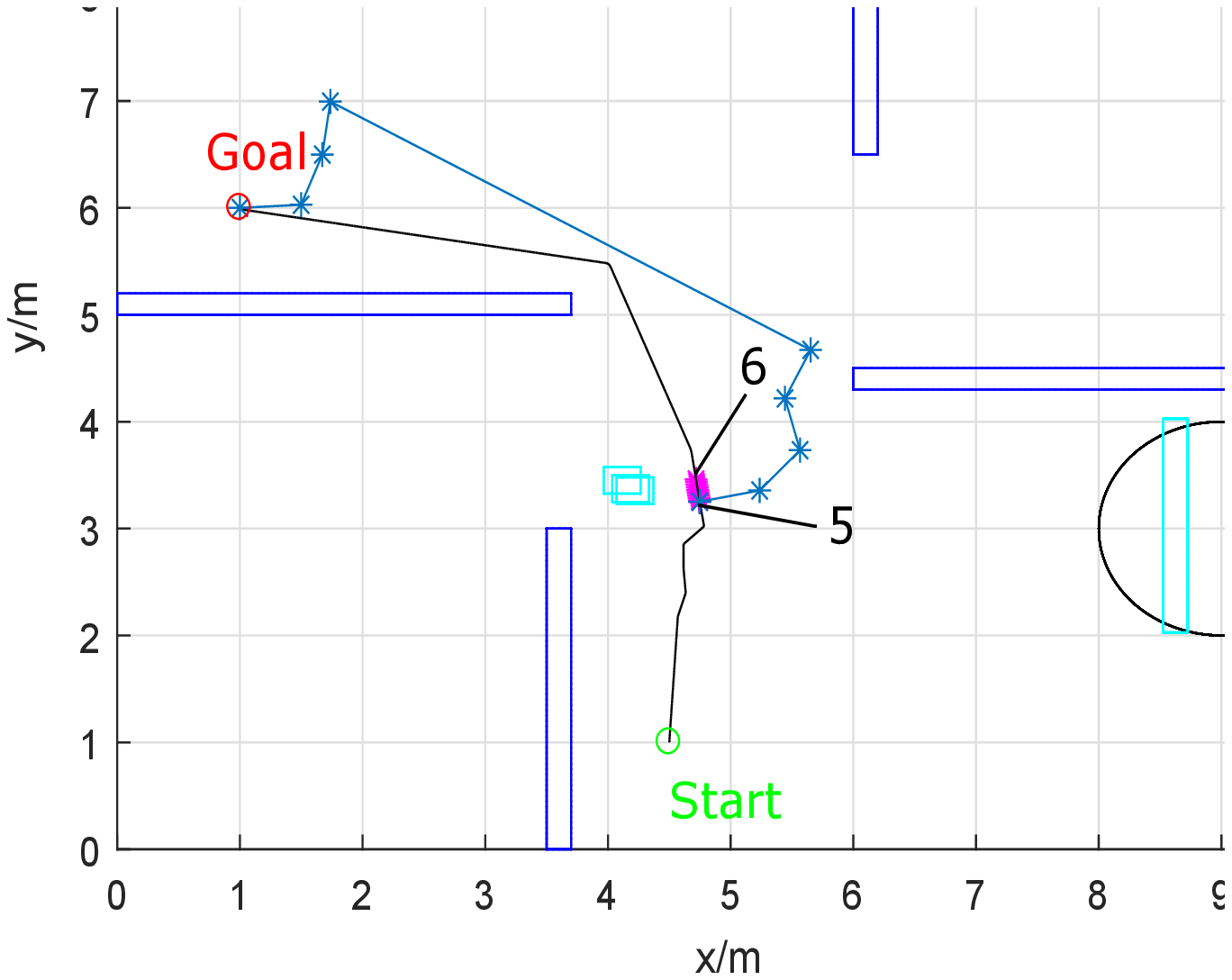,width=4cm}} 
	\centerline{{\scriptsize (d)}\psfig{file=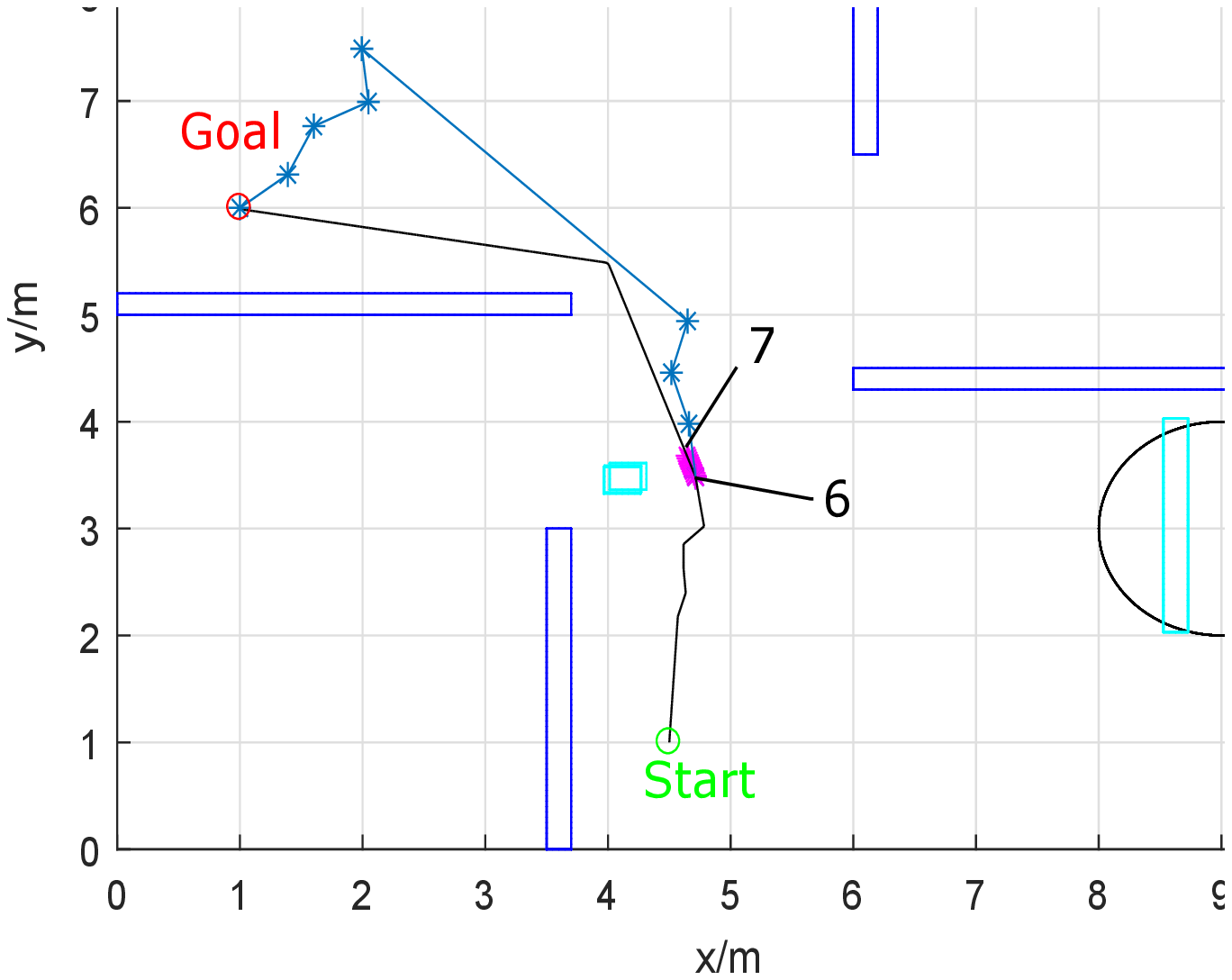,width=4cm}{\scriptsize (e)}\psfig{file=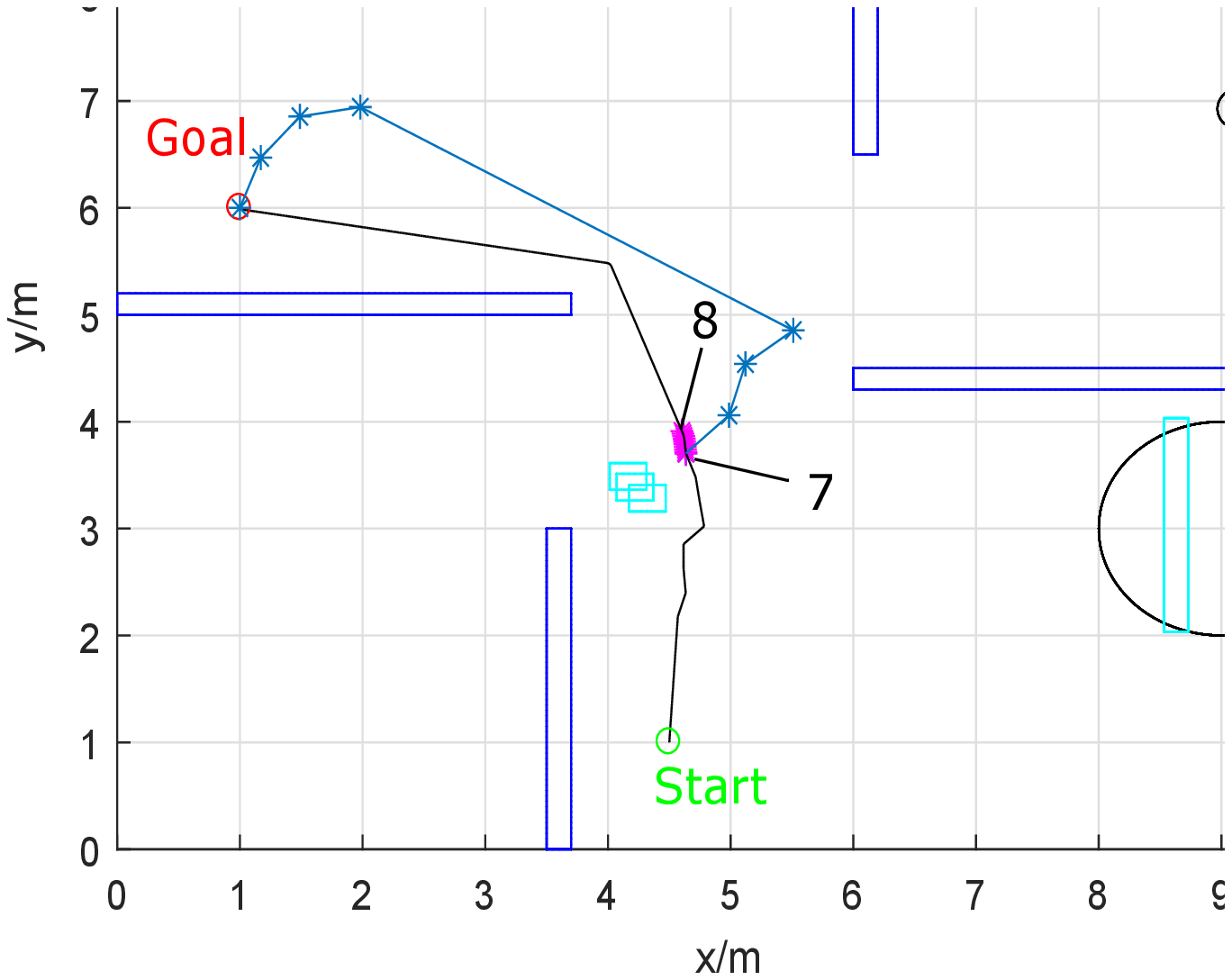,width=4cm}{\scriptsize (f)}\psfig{file=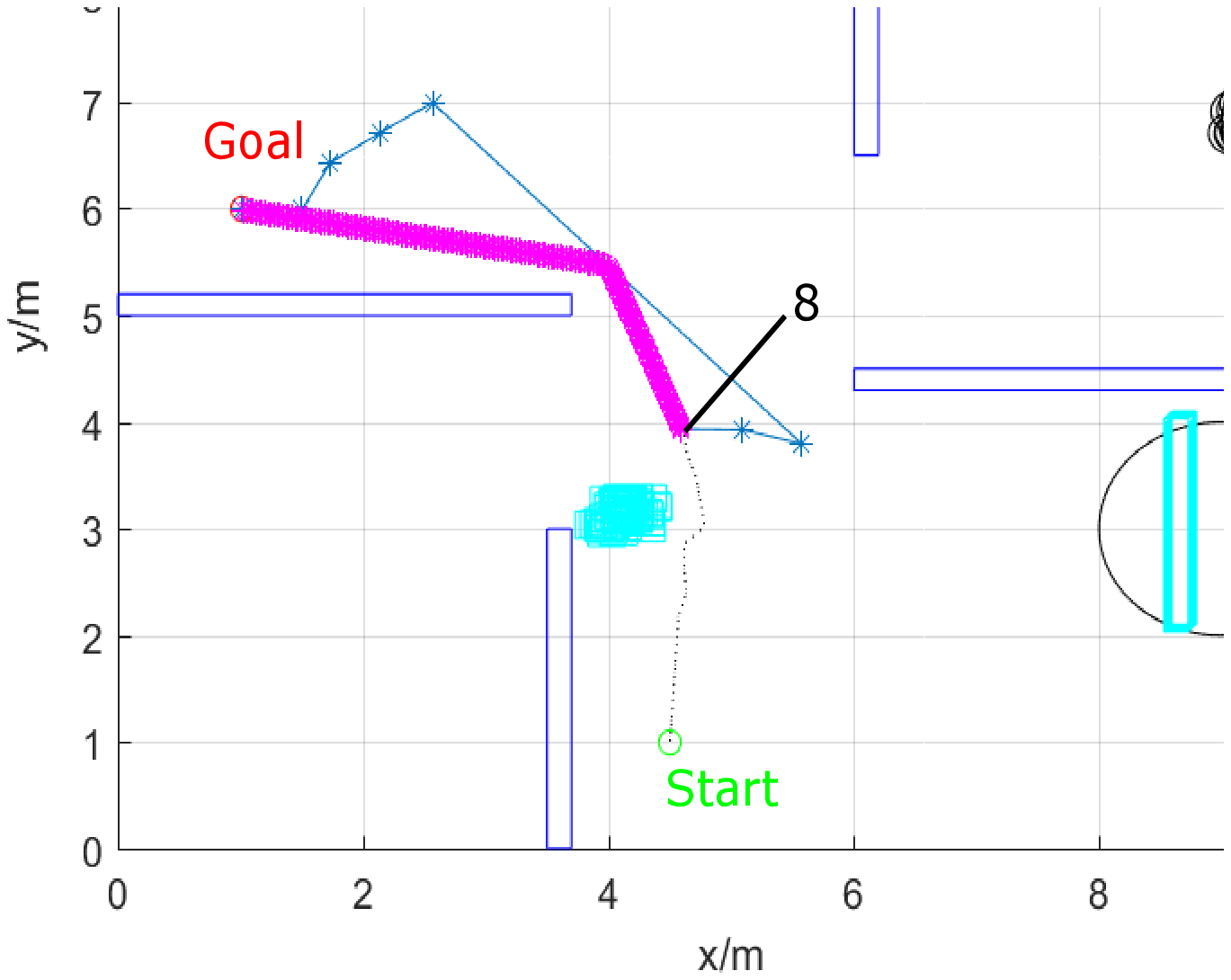,width=4cm}} 
	\caption{On-line motion planning results for  mobile base (II). (a) Phase $ 4 $, (b) Phase $ 5 $, (c) Phase $ 6 $, (d) Phase $ 7 $, (e) Phase $ 8 $, (f) Phase $ 9 $.}
	\label{fig:test2dd}
\end{figure}
Suppose that the mobile base locates initially at $(4.5,1)$  and its desired position is  $(1,6)$. 
Figs. \ref{fig:test2da} and \ref{fig:test2dd} show the motion planning results. 
In detail, Fig. \ref{fig:test2da} (a) shows the off-line motion planning results using \textbf{Algorithm \ref{alg:offline}}. The blue star points and line represent the non-optimal path. The red line represents the path after node rejection, and the green line represents the optimal path after node adjustment. The black dot  line represents the mobile base's trajectory using the linear polynomials with parabolic blends interpolation technique $ InterPolyBlend $. We can see that the trajectory
coincides with the optimal path. As a comparison, the magenta line represents the planned trajectory using B-spline interpolation which generates too many orientations for the mobile base and may bring unforeseen collisions. 

The star magenta points from Fig. \ref{fig:test2da} (b) to Fig. \ref{fig:test2dd} (f) represent the movements of mobile base. The black line behind the magenta points represents the mobile base's historical positions, and the black line ahead the magenta points represents the designed motion by \textbf{Algorithm \ref{alg:online}} in real time.  It can be seen that each time there are predicted collisions. The on-line planning process is activated  to design the new motion for mobile base.  
Throughout this simulation,  there are in total $ 8 $ \emph{On-linePlanning} calls,  which are illustrated by the blue star points and lines from phase $2$ to phase $9$, to update the motion  to avoid the cyan cuboid dynamic obstacle on the way to the desired position. Fig. \ref{fig:test2dd} (f) shows that  the motion remains the same after bypassing the dynamic cyan cuboid (see phase $9$).  
\begin{figure}[b]
	\centerline{{\scriptsize(a)}\psfig{file=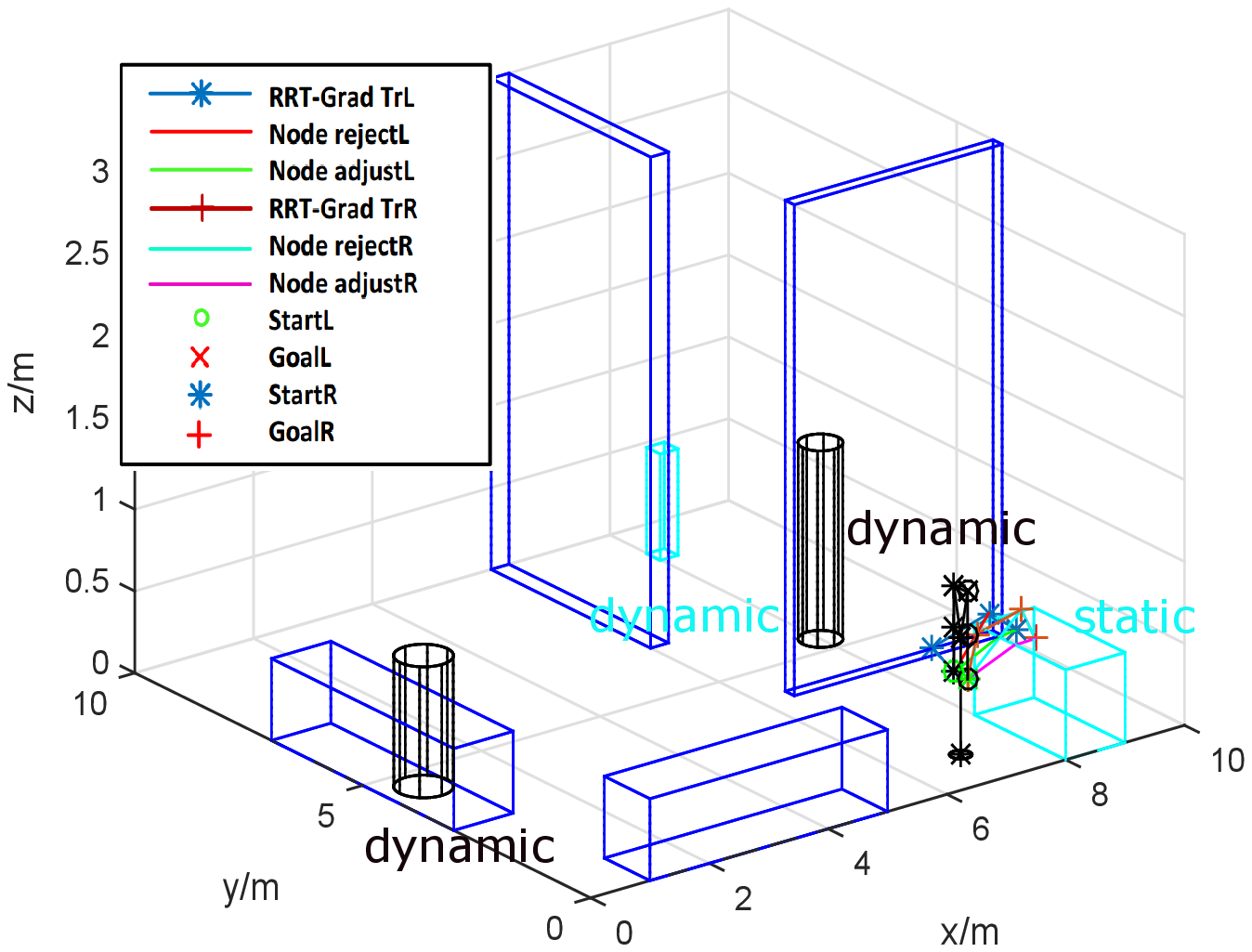,width=6cm}{\scriptsize (b)}\psfig{file=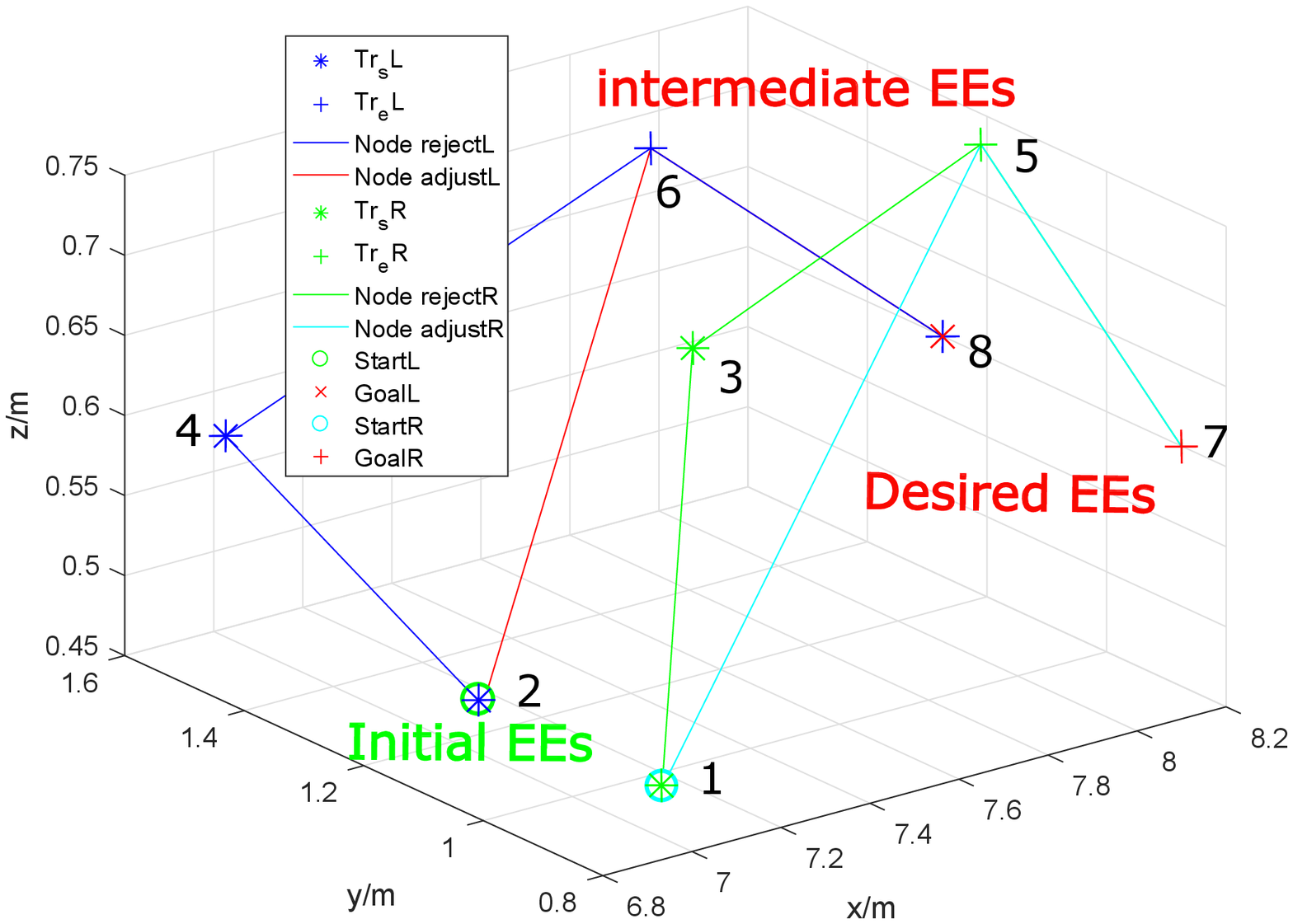,width=6cm}} 
	\centerline{{\scriptsize (c)}\psfig{file=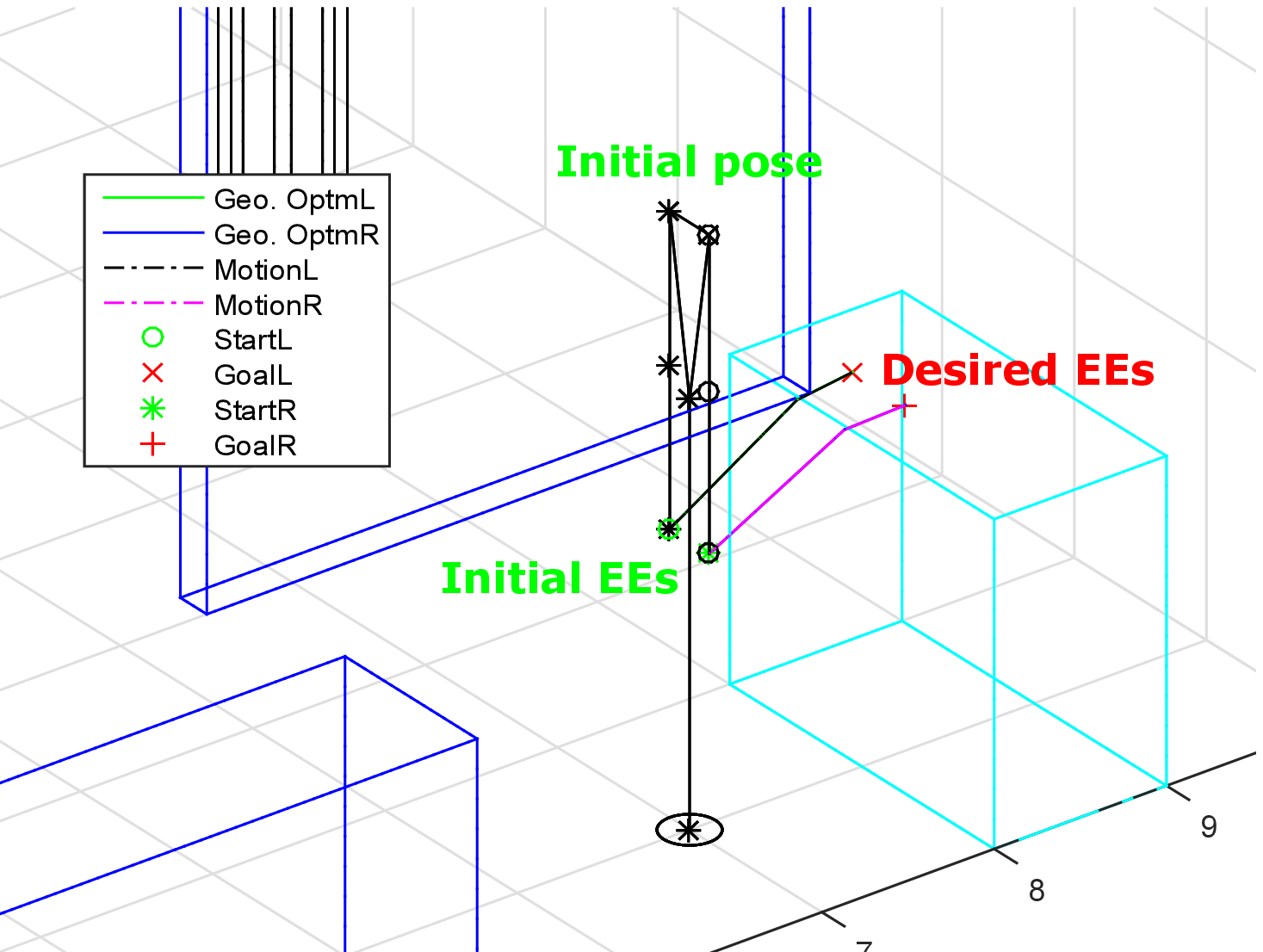,width=6cm}{\scriptsize (d)}\psfig{file=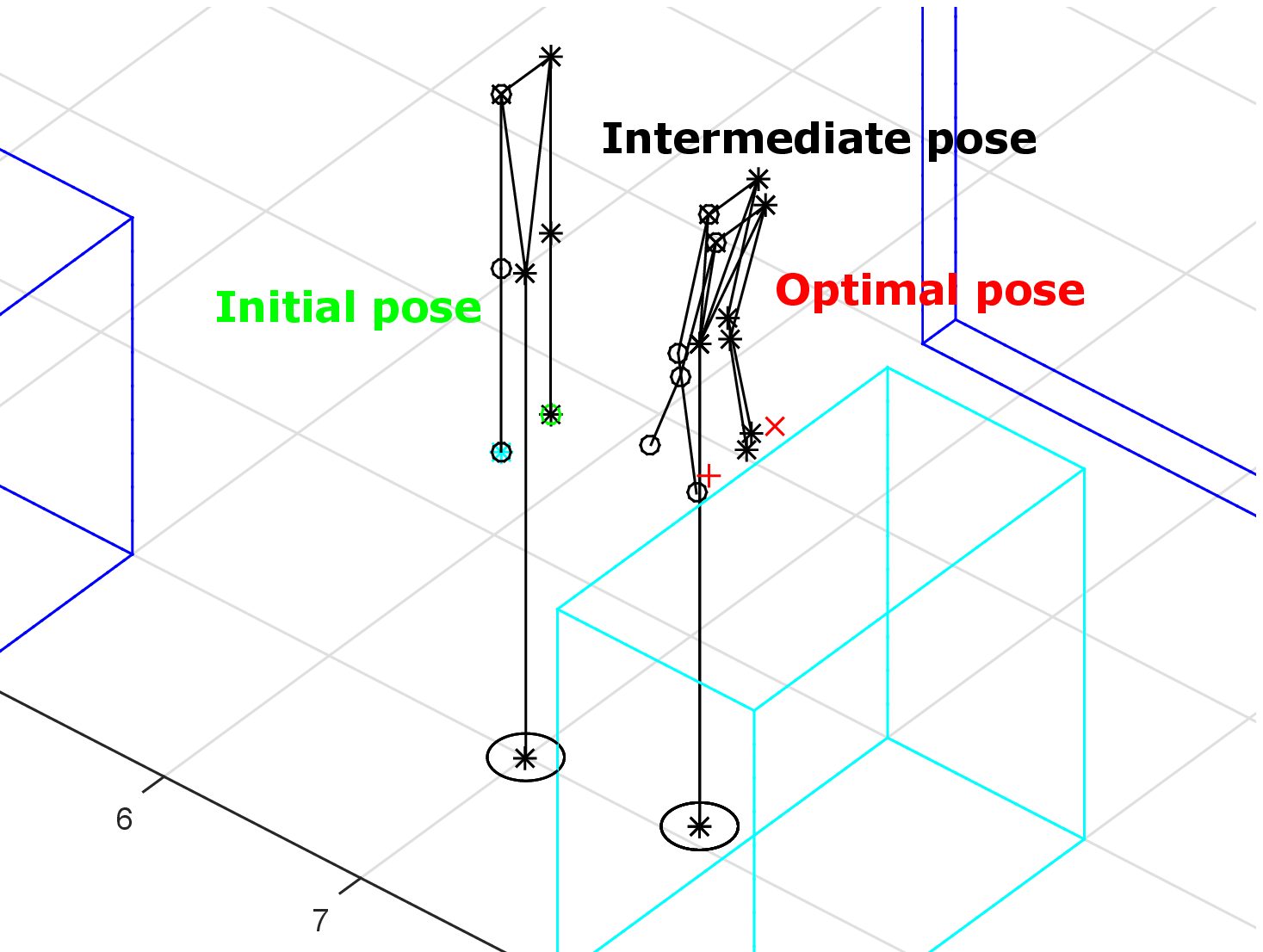,width=6cm}} 
	\caption{Motion planning results for MDAMS. (a) Path planning (global view) for EEs, (b) Path planning (local view) and via-points for EEs, (c) Trajectory planning for EEs, (d) Optimal via-poses for MDAMS.}
	\label{fig:test2ee}
\end{figure}

\subsection{Motion planning for MDAMS}
\noindent
Theoretically speaking, \textbf{Algorithm \ref{alg:online}} can be used to design the motion for any DoF mechanical systems, but the computational cost will increase and the via-poses will be difficult to predict.
This subsection is to verify the EEs' via-point-based MOGA motion planning algorithm.

Suppose that the initial pose of MDAMS is $\bm{\theta}_0 =(7,1,0,\bm{0}_{16})$; the left EE's  initial and desired positions are $(7,1.15,0.612)$ (green circle)  and  $(8.1,1.3,0.66)$ (see Fig. \ref{fig:test2ee} (a)), respectively over a $2m \times 1 m \times 0.6 m$  table (cyan cuboid) at $(9.5,1,0) $; the right EE's  initial and desired positions are $(7,0.85,0.612)$ (blue star) and  $(8.1,0.7,0.66)$  (red plus), respectively. 
The objective is  to reach  EEs' desired positions  among the obstacles with reasonable via-poses. 
The  motion planning results are shown in Fig. \ref{fig:test2ee}.

Firstly,  the EEs' via-points are  designed using \textbf{Algorithm \ref{alg:online}}. 
Fig. \ref{fig:test2ee} (a) shows the  path planning results for two EEs, 
Fig. \ref{fig:test2ee} (b) zooms in the results. There are in total three pairs of via-points, i.e. $1-2$, $5-6$ and $7-8$ in the optimal path. 
The pair $3-4$ is rejected after node adjustment. 
Fig. \ref{fig:test2ee} (c) shows EEs' trajectories bypassing  $1-2$, $5-6$ and $7-8$. However, the movements of  mobile base and other joints are not decided.
To this end, the EEs' via-point-based  MOGA algorithm is used to  optimize the  via-poses of the whole MDAMS. 
Fig. \ref{fig:test2ee} (d) shows the designed optimal  via-poses corresponding  to $1-2$, $5-6$ and $7-8$. It can be seen that the designed via-poses avoid the table and have the cognitive shock-free ``human-like'' behaviors. 

\section{Discussion}\label{sec:7}
\noindent
The main contributions of this paper are  ``human-like'' motion planning algorithms to improve the quality of human-robot interaction. They avoid calculating inverse Jacobian and satisfy  multiple constraints simultaneously. 

The  \textbf{Algorithm \ref{alg:MaxiMinNSGA}} is used to design simultaneously the mobile base's position-orientation  and the manipulator's configuration. Other constraints found in the literature can  be also integrated into  \textbf{Algorithm \ref{alg:MaxiMinNSGA}}, and it can be used for other multi-objective optimization problems.  
Theoretically speaking,  \textbf{Algorithm \ref{alg:MaxiMinNSGA}} can be used to design MDAMS's motion given the desired trajectories of EEs, but the shortcoming is that it is time-consuming. So it is not applicable for on-line motion planning.
To this end,  an efficient pose-to-pose bidirectional RRT and gradient decent motion planning algorithm with a geometric optimization method is proposed to design the motion from the initial pose to the optimal pose. 

Berenson$^{14,55}$ 
proved the probabilistic completeness of RRT-based algorithms under end-effector's orientation constraints. In this paper, the volume of the manifold in an embedding space is not zero, thus the probability of generating a path on the manifold of the proposed \textbf{Algorithm \ref{alg:offline}} by rejection sampling in the embedding space will go to $ 1 $ as the number of samples goes to infinity.

Geraerts$^{34}$ 
pointed out that many motion planning techniques generate low quality paths,  and presented a number of techniques to improve the quality of paths. 
However, complex geometric calculation of the medial axis is required and the effectiveness for  articulated robots remains to be studied. The same problem happened in {\color{blue} Wilmarth 89}. 
In this paper, only the planned via-points are needed to be adjusted within their neighborhoods to increase the clearance  without calculating the medial axis.
Compared with {\color{blue} Bakdi 17}, 
the proposed geometric optimization method  is simpler and more efficient without considering different cases in detail. 
Moreover, different from {\color{blue} Greiff 17} 
which uses a projection process to find the collision-free via-points, the \textbf{Algorithm \ref{alg:online}} in this paper can easily find them. The simple geometric optimization method can be applied to other planning methods like roadmaps or RRTs for path pruning. 
Different from the method in {\color{blue} Bakdi 17} 
which searches for a great number of via-poses from the initial pose to the optimal pose, the via-point-based  MOGA algorithm  in this paper only needs to design a few via-poses to achieve ``human-like'' collision-free behaviors. 


\section{Conclusion}\label{sec:6}
\noindent
The motion planning for a humanoid MDAMS is researched in this paper. 
The improved MaxiMin NSGA-II algorithm is proposed to design the optimal pose by optimizing five objective functions. 
Besides, an efficient  direct-connect bidirectional RRT and gradient descent  method is proposed to speed up greatly the sampling process.  And a geometric optimization method is designed to always guarantee the consistent and shortest collision-free path. ``Natural''  head forward behaviors are realized to ``tell'' motion intentions by assigning reasonable mobile base's orientations.
By designing on-line sensing, collision-test and control cycles, motion planning in dynamic environments with unknown obstacles is achieved. 
Furthermore, an EEs' via-point-based MOGA  motion planning algorithm is proposed to optimize the via-poses. 
The proposed  algorithms in this paper can be used for motion planning  for various robots. In order to implement the proposed motion methods and to improve the long-term performance of MDAMS, the future work will focus on  validation of the proposed algorithms on virtual prototype. Besides, uncertainties come from environments, sensors and computation will be taken into account for real applications.

\section*{References}
\myitem	H. S. Yang et al., Design and development of biped humanoid robot, AMI2, for social interaction with humans, in {\it IEEE-RAS Int. Conf. Humanoid Robots} (IEEE Press, Genova, Italy, 2006), pp. 352--357.


\myitem J. Lafaye, D. Gouaillier, P. B. Wieber, Linear model predictive control of the locomotion of Pepper, a humanoid robot with omnidirectional wheels, in {\it  IEEE-RAS Int. Conf.  Humanoid Robots} (IEEE Press, Madrid, Spain, 2014), pp. 336--341.

\myitem M. A. Diftler et al., Robonaut 2 - The first humanoid robot in space, in {\it  IEEE Int. Conf.  Robotics and Automation (ICRA), 2011}, pp. 2178--2183. 

\myitem Low, S. C.,  Phee, L., A review of master–slave robotic systems for surgery,  {\it International Journal of Humanoid Robotics (INT J HUM ROBOT)} {\bf 3}(04) (2006), 547--567.

\myitem S. Kuindersma et al., Optimization-based locomotion planning, estimation, and control design for the atlas humanoid robot, in {\it Autonomous Robot} {\bf 40}(3) (2016) 429--455. 

\myitem B. Cohen, S. Chitta, M. Likhachev, Search-based planning for dual-arm manipulation with upright orientation constraints, in {\it  IEEE Int. Conf.  Robotics and Automation (ICRA)}(IEEE Press, St. Paul, MN,  USA, 2012), pp. 3784--3790.

\myitem L. E. Kavraki et al., Probabilistic roadmaps for path planning in high-dimensional configuration spaces, {\it IEEE Trans. Robot. Autom.} {\bf 12}(4) (1996) 566--580. 

\myitem LaValle, Steven M., {\it Rapidly-exploring random trees: A new tool for path planning}[J]. 1998.

\myitem M. H. Korayem, S. R. Nekoo, The SDRE control of mobile base cooperative manipulators: Collision free path planning and moving obstacle avoidance, {\it Robotics and Autonomous Systems} {\bf 86}  (2016) 86--105.

\myitem R. Raja, A. Dutta, Path planning in dynamic environment for a rover using A* and potential field method, in {\it  Int. Conf. Advanced Robotics (ICAR)}(Hong Kong, China, 2017), pp. 578--582.

\myitem Y. C. Tsai, H. P. Huang, Motion planning of a dual-arm mobile robot in the configuration-time space, in {\it  IEEE/RSJ Int. Conf. Intelligent Robots and Systems (IROS)}(St. Louis (MO), USA, 2009), pp. 2458--2463.

\myitem C. L. Lewis, Trajectory generation for two robots cooperating to perform a task, in {\it  IEEE Int. Conf. Robotics and Automation(ICRA)} (IEEE Press, Minnesota, USA, 1996),  vol.2., pp. 1626--1631.

\myitem J. P. Saut et al.,  Planning pick-and-place tasks with two-hand regrasping, in {\it  IEEE/RSJ Int. Conf. Intelligent Robots and Systems (IROS)}(IEEE Press, Taipei, Taiwan, 2010), pp. 4528--4533. 

\myitem D. Berenson, S. S. Srinivasaz, Probabilistically complete planning with end-effector pose constraints, in {\it  IEEE Int. Conf. Robotics and Automation (ICRA)}(IEEE Press, Alaska,  USA, 2010), pp. 2724--2730.

\myitem P. Michel  et al., Motion planning using predicted perceptive capability, {\it International Journal of Humanoid Robotics}, {\bf 6}(03) (2009), 435-457. 

\myitem M. H. Korayem, M. Nazemizadeh, V. Azimirad, Optimal trajectory planning of wheeled mobile manipulators in cluttered environments using potential functions, {\it Scientia Iranica} {\bf 18}(5) (2011) 1138--1147.

\myitem M. H. Korayem, V. Azimirad, M. I. Rahagi, Maximum Allowable Load of Mobile Manipulator in the Presence of Obstacle Using Non-Linear Open and Closed Loop Optimal Control, {\it Arabian Journal for Science and Engineering} (AJSE) {\bf 39}(5) (2014) 4103--4117.

\myitem K. Harada et al.,  Base position planning for dual-arm mobile manipulators performing a sequence of pick-and-place tasks, in {\it  IEEE-RAS 15th Int. Conf. Humanoid Robots (Humanoids)} (IEEE Press, Seoul, South Korea, 2015), pp. 194--201. 

\myitem M. Galicki, Real-time constrained trajectory generation of mobile manipulators, {\it Robotics and Autonomous Systems} {\bf 78} (2016) 49--62.

\myitem K. Deb, {\it Multi-objective optimization using evolutionary algorithms}, vol.16, (NY, USA, John Wiley \& Sons, 2001).

\myitem T. T. Mac et al.,  Heuristic approaches in robot path planning: A survey, {\it Robotics and Autonomous Systems} {\bf 86}(Supplement C) (2016) 13--28. 

\myitem J. K. Parker, A. R. Khoogar, D. E. Goldberg, Inverse kinematics of redundant robots using genetic algorithms, in {\it Int. Conf. Robotics and Automation  (ICRA)}(IEEE Press, Arizona, USA, 1989), pp. 271--276 vol.1.

\myitem M. Zhao, N. Ansari, E. S. H. Hou, Mobile Manipulator Path Planning By A Genetic algorithm, in {\it Proceedings of the IEEE/RSJ Int. Conf. Intelligent Robots and Systems (IROS'92)}(North Carolina, USA, 1992), vol.1, pp. 681--688.

\myitem M. d. G. Marcos, J. A. Tenreiro Machado, T. P. Azevedo-Perdicolis, A multi-objective approach for the motion planning of redundant manipulators,{\it Applied Soft Computing} {\bf 12}(2) (2012) 589--599.

\myitem A. Bakdi et al.,  Optimal path planning and execution for mobile robots using genetic algorithm and adaptive fuzzy-logic control, {\it Robotics and Autonomous Systems} {\bf 89} (2017) 95--109. 

\myitem D. E. Goldberg, Genetic Algorithms in Search, Optimization and Machine Learning, 1st edn. (Boston, MA, USA, Addison-Wesley Longman Publishing Co., Inc.,  1989). 

\myitem N. Srinivas, K. Deb, Muiltiobjective Optimization Using Nondominated Sorting in Genetic Algorithms, {\it Evolutionary Computation} {\bf 2} (1994) 221--248.

\myitem K. Deb et al.,  A fast and elitist multiobjective genetic algorithm: NSGA-II, {\it IEEE Trans. Evol. Comput.} {\bf 6}(2) (2002) 182--197. 

\myitem E. J. S. Pires, P. B. d. M. Oliveira, J. A. T. Machado, Multi-objective MaxiMin Sorting Scheme, in {\it International Conference on Evolutionary Multi-Criterion Optimization}  (Springer, Berlin, Heidelberg, 2005), pp. 165--175.

\myitem Y. Choi, H. Jimenez, D. N. Mavris, Two-layer obstacle collision avoidance with machine learning for more energy-effcient unmanned aircraft trajectories, {\it Robotics and Autonomous Systems} {\bf 98}(Supplement C) (2017) 158--173.

\myitem O. Adiyatov, H. A. Varol, A novel RRT*-based algorithm for motion planning in Dynamic environments, in {\it  IEEE Int. Conf. Mechatronics and Automation (ICMA)}(Takamatsu, Japan, 2017), pp. 1416--1421. 

\myitem S. Primatesta, L. O. Russo, B. Bona, Dynamic trajectory planning for mobile robot navigation in crowded environments, in {\it  IEEE 21st Int. Conf. Emerging Technologies and Factory Automation (ETFA)}(Berlin, Germany, 2016), pp. 1--8.

\myitem S. A. Wilmarth, N. M. Amato, P. F. Stiller, MAPRM: a probabilistic roadmap planner with sampling on the medial axis of the free space, in {\it Proceedings 1999 IEEE Int. Conf. Robotics and Automation (ICRA)},  pp. 1024--1031 vol.2. 

\myitem R. Geraerts, M. H. Overmars, Clearance based path optimization for motion planning, in {\it  IEEE Int. Conf. Robotics and Automation (ICRA)}(LA, USA,2004),  pp. 2386--2392 Vol.3. 

\myitem T. Mercy et al.,  Spline-Based Motion Planning for Autonomous Guided Vehicles in a Dynamic Environment, {\it IEEE Trans. Control Syst. Technol.} (2017) pp. 1--8. 

\myitem S. Lee, H. Moradi, C. Yi, A real-time dual-arm collision avoidance algorithm for assembly, in {\it  IEEE International Symposium on Assembly and Task Planning (ISATP)}(California, USA, 1997), pp. 7--12.

\myitem J. P. van den Berg, M. H. Overmars, Roadmap-based motion planning in dynamic environments, {\it IEEE Trans. Robot.} {\bf 21}(5) (2005) 885--897.

\myitem G. Pajak, I. Pajak, Motion planning for mobile surgery assistant, {\it Acta Bioeng Biomech} {\bf 16}(2) (2014) 11--20.

\myitem Y. Yang, V. Ivan, S. Vijayakumar, Real-time motion adaptation using relative distance space representation, in {\it 2015 Int. Conf. Advanced Robotics (ICAR)}, pp. 21--27.

\myitem G. Chen et al.,  A novel autonomous obstacle avoidance path planning method for manipulator in joint space, in {\it 2014 IEEE Conference on Industrial Electronics and Applications}, pp. 1877--1882. 

\myitem G. Xin et al.,  Real-time dynamic system to path tracking and collision avoidance for redundant robotic arms, {\it The Journal of China Universities of Posts and Telecommunications} {\bf 23}(1) (2016) 73--96. 

\myitem D. Han, H. Nie, J. Chen, M. Chen, Dynamic obstacle avoidance for manipulators using distance calculation and discrete detection, {\it Robotics and Computer-Integrated Manufacturing} {\bf 49}(Supplement C) (2018) 98--104.

\myitem P. D. H. Nguyen et al.,  A fast heuristic Cartesian space motion planning algorithm for many-DoF robotic manipulators in dynamic environments, in {\it IEEE-RAS 16th Int. Conf. Humanoid Robots (Humanoids)}(Cancun, Mexico, 2016), pp. 884--891. 

\myitem J. Vannoy, J. Xiao, Real-Time Adaptive Motion Planning (RAMP) of Mobile Manipulators in Dynamic Environments With Unforeseen Changes, {\it IEEE Trans. Robot.} {\bf 24}(5) (2008) 1199--1212.

\myitem Zhao, J., Wei, Y. (2017), A Novel Algorithm of Human-Like Motion Planning for Robotic Arms, {\it International Journal of Humanoid Robotics} {\bf 14}(01) (2017), 1650023.

\myitem Wu, Q. C., Wang, X. S.,  Du, F. P.,  Analytical inverse kinematic resolution of a redundant exoskeleton for upper-limb rehabilitation, {\it International Journal of Humanoid Robotics} {\bf 13}(03) (2016) 1550042.

\myitem C. Lamperti, A. M. Zanchettin, P. Rocco, A redundancy resolution method for an anthropomorphic dual-arm manipulator based on a musculoskeletal criterion, in {\it  IEEE/RSJ Int. Conf. Intelligent Robots and Systems (IROS)} (Hamburg, Germany, 2015), pp. 1846--1851.

\myitem S. Nishiguchi et al.,  Theatrical approach: Designing human-like behaviour in humanoid robots,{\it Robotics and Autonomous Systems} {\bf 89} (2017) 158--166. 

\myitem M. M. Emamzadeh, N. Sadati, W. A. Gruver, Fuzzy-based interaction prediction approach for hierarchical control of large-scale systems, {\it Fuzzy Sets and Systems} {\bf 329}(Supplement C) (2017) 127--152.

\myitem Y. Qian, {\it Design and Control of a Personal Assistant Robot}, Ph.D. thesis (2013).

\myitem J. H. Holland, Adaptation in natural and artificial systems: an introductory analysis with applications to biology, control, and artificial intelligence, (MIT press, 1992).

\myitem B. Xian et al.,  Task-Space Tracking Control of Robot Manipulators via Quaternion Feedback, {\it IEEE Trans. Robot. Autom.} {\bf 20}(1) (2004) 160--167. 

\myitem L. Sciavicco, B. Siciliano, {\it Modelling and Control of Robot Manipulators, Advanced textbooks in control and signal processing}, 2nd edn. (London, Springer, 2005).

\myitem S. McLeod, J. Xiao, Real-time adaptive non-holonomic motion planning in unforeseen dynamic environments, in {\it  IEEE/RSJ Int. Conf. Intelligent Robots and Systems (IROS)}(Daejeon, South Korea, 2016), pp. 4692--4699.

\myitem D. Berenson, S. Srinivasa, J. Kuffner, Task Space Regions: A framework for pose-constrained manipulation planning, {\it The International Journal of Robotics Research} {\bf 30}(12) (2011) 1435--1460.

\myitem M. Greiff, A. Robertsson, Optimisation-based motion planning with obstacles and priorities, {\it IFAC-PapersOnLine} {\bf 50}(1) (2017) 11670--11676.


\eject

\end{document}